\documentclass{article}

\usepackage{arxiv}

\usepackage[utf8]{inputenc} 
\usepackage[T1]{fontenc}    
\usepackage{hyperref}       
\usepackage{url}            
\usepackage{booktabs}       
\usepackage{amsfonts}       
\usepackage{nicefrac}       
\usepackage{microtype}      
\usepackage{graphicx}
\usepackage{xcolor}
\usepackage[linesnumbered,ruled,vlined]{algorithm2e}
\usepackage{amsthm,amssymb,latexsym, subfig}
\usepackage{amsmath}
\newcommand{\TEXTFONTT}[1]{\mathcal{#1}}
\newcommand{\tX}{\TEXTFONTT{X}}

\newcommand{\R}{\mathbb{R}}

\makeatletter
\newcommand*{\centerfloat}{%
  \parindent \z@
  \leftskip \z@ \@plus 1fil \@minus \textwidth
  \rightskip\leftskip
  \parfillskip \z@skip}
\makeatother

\title{Coarse-Grain Cluster Analysis of Tensors With Application to Climate Biome Identification}

\author{
 Derek DeSantis \\
  Los Alamos National Laboratory,\\
  Center for Nonlinear Studies\\
  \texttt{ddesantis@lanl.gov} \\
   \And
 Phillip J. Wolfram \\
  Los Alamos National Laboratory\\
  \texttt{pwolfram@lanl.gov} \\
  \And
 Katrina Bennett\\
Los Alamos National Laboratory\\
  \texttt{kbennett@lanl.gov} \\
  \And
   Boian Alexandrov\\
Los Alamos National Laboratory\\
  \texttt{boian@lanl.gov} \\
}

\begin{document}
\maketitle

\vspace{10pt}

\begin{abstract}

A tensor provides a concise way to codify the interdependence of  complex data.  Treating a tensor as a d-way array, each entry records the interaction between the different indices.    Clustering provides a way to parse the complexity of the data into more readily understandable information. Clustering methods are heavily dependent on the algorithm of choice, as well as the chosen hyperparameters of the algorithm.  However, their sensitivity to data scales is largely unknown.

In this work, we apply the discrete wavelet transform to analyze the effects of coarse-graining on clustering tensor data. We are particularly interested in understanding how scale effects clustering of the Earth's climate system. The discrete wavelet transform allows classification of the Earth's climate across a multitude of spatial-temporal scales.  The discrete wavelet transform is used to produce an ensemble of classification estimates, as opposed to a single classification. Information theoretic approaches are used to identify important scale lenghts in clustering The L15 Climate Datset. We also discover a sub-collection of the ensemble that spans the majority of the variance observed, allowing for efficient consensus clustering techniques that can be used to identify climate biomes. 

\end{abstract}


\section{Introduction}


Data measured from a high-order complex system can be difficult to analyze.  A convenient tool to store this data is in the form of a tensor, or d-way array. Each entry of the array describes the value obtained across the $d$ parameters. Often, the  dependencies between indices is not clear, making direct interpretation of the data a demanding task.

Numerous methods have been developed to allow one to readily parse these complex interdependencies to provide meaningful interpretations of the data.  Among the most well studied methods are tensor factorizations and clustering techniques.  Different forms of tensor factorizations have been shown to be effective at multi-way dimensionality reduction, blind source separation, data mining, and latent feature extraction \cite{cichocki2009nonnegative, kolda2009tensor}. Indeed, nonnegative tensor factorizations, a constrained tensor factorization requiring the data and each factorized component to consist of nonnegative real numbers,  has had success at extracting hidden, interpretive features of the data \cite{alexandrov2019nonnegative, lopez2019unsupervised, schein2016bayesian, stanev2018unsupervised, vesselinov2019unsupervised}.  This is because many types of real-world data are naturally non-negative, so the enforced positivity constraint on the factors can be interpreted as a mixing of physical signals \cite{lee1999learning}. 

While the goal of factorization techniques are often to discover latent structures within the data, the aim of clustering is to provide coherent groupings of objects. When data comes in the form of vectors (order-1 tensors), there is an immense, ever growing lexicon of clustering methods \cite{fahad2014survey}. For clustering general tensors there is a larger flexibility of techniques due to the interactions along different  fibers of the tensors.  This leads to more complex clustering problems, requiring new optimization algorithms to approximately solve for the desired clustering \cite{cao2013robust, jegelka2009approximation}.  

These two goals of factorization and clustering are often deeply related.  For example, it has been shown that by adding constraints to nonnegative matrix factorization (order-2 tensor factorization), one can produce an optimization function identically to K-means or spectral clustering (order-1 clustering) \cite{ding2008equivalence}.  Similarly, there is an equivalence between higher order singular value decomposition and a K-means clustering for tensors with additional constraints \cite{huang2008simultaneous}. 

\subsection{Classifying climate biomes}


We are interested in leveraging the unsupervised learning techniques discussed above to assist in the interpretation of climate data.  Climate data generally arises as a collection of spatial-temporal measurements of various physical and biological features of The Earth system (e.g. temperature, precipitation, flora and fauna). The tensor of climate data compactly records the complex interdependence between space and time for different variables of interest.  Recently, tensor factorization techniques have found success at extracting latent climate signals.  For example, see \cite{alexandrov2014blind,zhang2009parallel}. Clustering the climate into locally similar spaces is a classical problem with a longer history.

A \textit{biome} is a region of space that has homogeneous climate features, e.g., similar temperature profiles. The process of finding climate biomes is clustering, and a biome is a synonym for a cluster of spatial points on the Earth. Throughout this text, we use the word biome and cluster interchangeably.
Historically, the standard to classify climate biomes has been the K\"oppen-Geiger (KG) model \cite{kottek2006world}.  The KG model is an expert-based judgment that describes climate zones using temperature and precipitation measurements.   The KG model utilizes a fixed, expert opinion-based decision tree, where each branch uses various information about temperature and precipitation. This heuristic allows one to broadly assess climate regions.  

While KG is interpretable, it is overly simplistic and somewhat arbitrary.  In an attempt to remedy this problem, Thornthwaite \cite{thornthwaite1948approach} introduced a more nuanced model using moisture and thermal factors.  However, the Thornthwaite model (along with its successors) still suffer from expertly chosen biases in their parameters.  

A solution to this problem is to move towards data-driven methods of classification. Here,  human bias is subsummed onto the machine learning algorithm  that seeks to minimize some cost function.  This is equivalent to a statistical assumption about the data generation and distribution \cite{bishop2006pattern}.  In the views of the authors, this is often a more reasonable assignment of bias. In \cite{zscheischler2012climate}, Zscheischler et. al.  compare KG to the K-means algorithm. They show that, unsurprisingly, K-means outperforms KG with respect to statistical measures, specifically explained predictand variance and variation of information. In \cite{netzel2016using}, the authors use mean monthly climate data to perform hierarchical clustering and partition around medoids.  In each clustering algorithm, two distance metrics are tested, and these results are compared to KG using an information-theoretic measure.

These data driven approaches to climate clustering are an epistemological improvement over the user chosen heuristics of KG.  However, these clustering methods suffer from challenges external to the climate science application space. 

\subsection{Clustering challenges}


Abstractly, a  \textit{clustering} is any function that takes a dataset $X=\{x_n\}_{n=1}^N$ and returns a partitioning of $X$. One usually seeks a clustering that satisfies a chosen heuristic.  For example, a K-means clustering is any partition of $X$ into disjoint non-empty sets $U_1, \dots U_K$ that minimizes inner cluster variance.  Namely, 
       \begin{equation}
       \label{kmeanseq}
        \mbox{Argmin}_{U_1, \dots U_K} \sum_{k=1}^K \sum_{x_n \in U_k} \| x_n - m_k\|^2
        \end{equation}
where $m_k$ is the mean of $U_k$. The number of possible ways to put the $N$ data points into $K$ clusters is ${N-1\choose K-1}$. For $N \gg K$, it becomes infeasible to search for the optimal K-means clustering directly.  Indeed, finding the optimal solution to Equation \ref{kmeanseq} is NP-hard \cite{mahajan2009planar}.  As a result, optimization schemes such as Lllyod's algorithm have been developed to approximate the optimal solution.  These algorithms are necessarily nondeterminstic, and therefore may fail to adequately approximate the global minimum if the data isn't sufficiently nice, e.g. low signal to noise ratio \cite{lu2016statistical}.  

The above example of K-means summarizes an issue persistent across many approaches to clustering.  A heuristic is chosen to group the data, but unfortunately finding the cluster(s) that satisfy this heuristic is computationally intractable.  As a result, algorithms are developed to efficiently compute an approximation to these optimal solutions.  Because these clustering methods cannot guarantee convergence to the optimal solution for all datasets, different clustering measures have been formulated to assess the quality of a clustering. However, often the measures are directly exported from the optimization functions used in the clustering algorithm.  The algorithm that is designed to optimize this clustering measure will, by design, out perform other clustering methods with respect to that metric. As a result, this provides no further information as to what clustering strategy is better suited for the problem.

These challenges highlight that there is no true ``best'' clustering in general.   Rather, there are many ideal clusterings that arise from the specifics of the scientific inquiry pursued.  None of the optimal clusterings is certifiably ``correct'', but each provides different insights into the structure of the data. Unfortunately, the quality of the obtained clustering is not easy to evaluate.

These problems have led researchers to define the concept of an ensemble, or consensus clustering \cite{nguyen2007consensus}. Here, many clusterings are combined to produce a single clustering of the data.  Common features between the clusterings are amplified, and artifacts become dulled (Figure \ref{fig:consensus}a). There is evidence to suggest that selecting a smaller ensemble with good, diverse clusters outperforms larger, redundant ensembles \cite{caruana2004ensemble, fern2008cluster, hadjitodorov2006moderate, kuncheva2004using}. For this reason, we believe it is preferable to find a smaller ensemble of quality, diverse clusters. 

\begin{center}
\centering
\begin{figure}[h]
\centerfloat
    \begin{tabular}{cc}
        \subfloat[Consensus Clustering:  Each clustering arises from choice of different clustering algorithm or different hyper-parameters.]{\includegraphics[width = 0.45\textwidth]{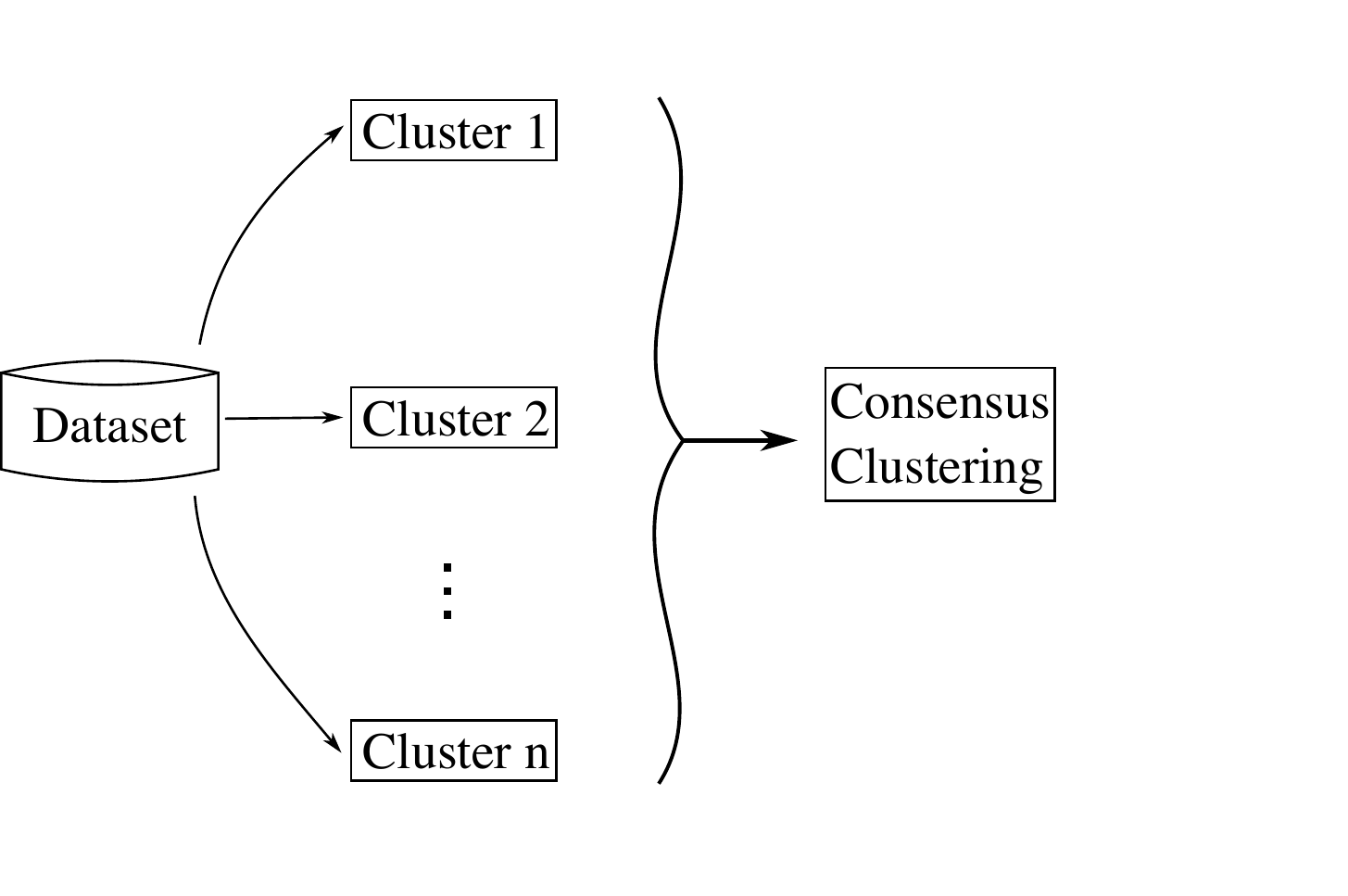}} &
        
        \subfloat[Consensus Clustering + CGC \& MIER: From each clustering, we propose to run CGC and the MIER algorithm.  This results in a larger ensemble that will minimally capture the dependence of the data on scale.]{\includegraphics[width = 0.45\textwidth]{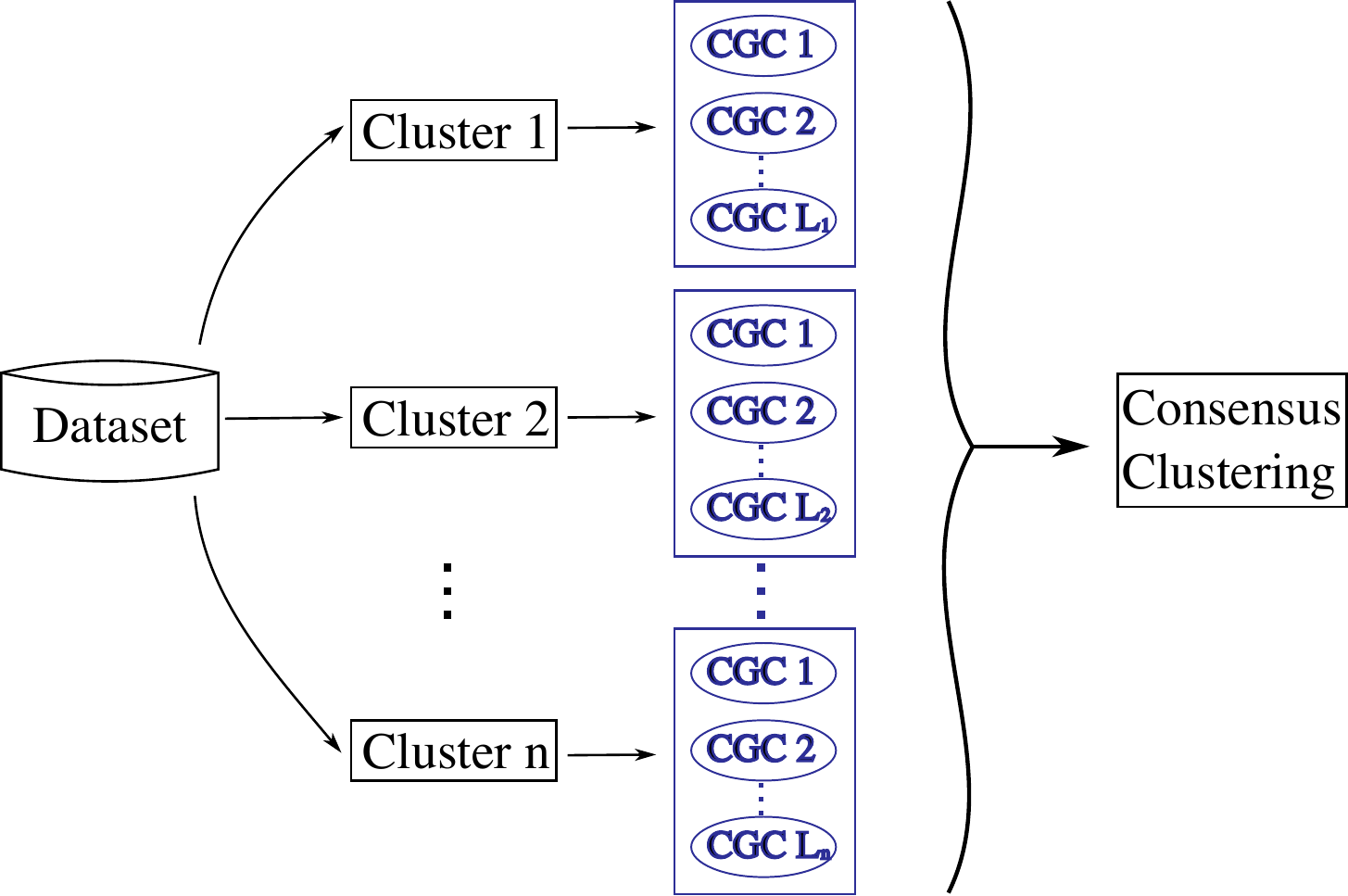}}

    \end{tabular}
    \caption{Proposed augmentation to the consensus clustering workflow.}
    \label{fig:consensus}
\end{figure}
\end{center}

\subsection{Our Contributions}


There are still unresolved issues not addressed by the current ensemble clustering framework.  The choice of a heuristic for any single clustering is a proxy for a potentially ill-specified human objective. This ignores other aspects of the environment, weighting an indifference to other potentially important environmental variables \cite{amodei2016concrete}.  Consequently when a user can identify potential problems with their ML architecture, they need to resolve the effect the issue has on the overall result.

One practical example in the climate sciences is the scale at which the data is acquired. Most natural or environmental data is formed by directly observing and measuring quantities where the underlying or driving processes are usually unknown. The hidden or latent features of the data may not clearly present themselves at the resolution that the data was sampled.  For example, weather data is often gathered at fine temporal and spatial detail, e.g., daily temperature at a single weather station. However, climate signals are often observed on the order of years or decades and across a region.

\newpage

The above discussion highlights two important problems:
\begin{enumerate}
    \item[1.] The need to address hidden parameters effecting the data, such as scale dependence
    \item[2.] The desire to build an ensemble of clusters for a consensus clustering algorithm.
\end{enumerate}
\textbf{In this work, we discuss a clustering method that tackles both of these problems.} 
We develop a technique that uses the discrete wavelet transform to cluster slices of tensors at different scales  that we call \textit{coarse-grain clustering} (CGC).  
This highly parallelizable technique results in many potential clusterings, one for each chosen coarse-graining.  
Not all of these coarse-grain clusterings provide new information, however.   
Thus, we present a novel selection method that leverages mutual information between clusterings to quantify the loss of information between clusterings and select a small subset that best represents ensemble. 
We call this reduction algorithm \textit{Mutual Information Ensemble Reduce} (MIER).

While the end-to-end workflow we have discussed involves ideas from traditional consensus clustering (e.g., Figure \ref{fig:consensus}a), the focus of this paper is specifically on a novel modification to this approach leveraging the CGC and the MIER algorithms to develop a classification using coarse-grain clustering (CGC) and in accordance with a mutual information ensemble reduction (MIER), i.e., the blue highlighted portion of Figure \ref{fig:consensus}b.

This paper is organized as follows.  First background material used for development of the CGC and MIER algorithms is presented in Section 2. 
The structure of these algorithms is detailed in Section 3.  In Section 4, the algorithm is applied to a widely-used climate data set as a case study with presentation of results and discussion, followed by conclusions and a recommendation for future work in Section 5.

\section{Preliminaries}

In this section, we briefly review key mathematical tools used throughout this work including 1) the discrete wavelet transform and its role in separating earth system data into spatio-temporal scales, 2)  graph cuts and their connection to spectral clustering, and 3) use of mutual information to measure similarity between two clusters of the same data. 

\subsection{Discrete Wavelet Transform (DWT)}

Given a one-dimensional function $f: \mathbb{N} \rightarrow \R$ the \textit{discrete wavelet transform} (DWT) is a process of iteratively decomposing $f$ into a series of low and high frequency signals. This process is accomplished by convolving the function $f$ with low frequency and high frequency filter functions that arise from a choice of \textit{mother wavelet} function $\Psi$, sometimes called a wavelet for short.  The low frequency signal is often referred to as the \textit{approximation coefficients}, and the high frequency is called the \textit{detail coefficients}.  The filtering process removes half the signals frequencies, so downsamping can then be performed without reconstruction loss on $f$. As a result, if the signal $f$ consists of $N$ data points, then the low and high frequency signals after the DWT contain approximately $\frac{N}{2}$ data points.  See the Appendix for more details. 

Fixing a scale level $\ell \in \mathbb{N}$, the signal $f$ is decomposed into its high and low frequency signals $\ell$ times.  The discrete wavelet transform plays an important role in our work.  The DWT is useful to our analysis for the following reasons:
\begin{enumerate}
    \item By filtering data down to different scale lengths, clustering the approximation coefficients allows us to directly compare how the scale of the data effects the clustering result.  For our particular application, the approximation coefficients have an interpretation in terms of the analysis. Coarse-graining temporal signals captures seasonal, yearly, and eventually decadal trends, whereas coarse-graining spatial information captures city-, county-, and eventually state-sized features.  Therefore, a comparison of clusterings from different coarse-grainings allows one to parse how these different scales effect clustering.  
    \item The DWT runs fast. The filter bank implementation of the DWT runs in $\mathcal{O}(N)$ for each 1-D implementation \cite{shukla2013efficient}, or $\mathcal{O}(N_1 N_2 N_3)$ for 3-D tensors.  Furthermore, we are only interested in the approximation coefficients, which have drastic size savings, as discussed above. 
    \item The coarse coefficients at the end of the filter bank naturally creates a hierarchy of datasets that, when clustered, provide an ensemble for a consensus clustering algorithm. 
\end{enumerate}

\subsection{Clustering algorithms and graph cuts}


As discussed in the introduction, we are interested in analyzing how the scale of data effects clustering.  In this subsection, we briefly discuss the unsupervised learning tools we use throughout the paper.  We outline how they work, and discuss emphasize our choice for use.

\subsubsection{K-means and spectral clustering}

Clustering algorithms are diverse with varying advantages and disadvantages \cite{fahad2014survey}.  Arguably the most famous are partitioned based algorithms, where data are iteratively reassigned to clusters until an optimization function is minimized.  
The prototypical example of a partitioned based clustering algorithm is \textit{K-means} discussed in the introduction. Given a natural number $K$, the K-means algorithm seeks to partition the data-set into $K$ distinct groups that minimize the variance within the clusters.  Lloyd's algorithm, the implementation of K-means we use in this work, is well known to run in $\mathcal{O}(NDKI)$, where $N$ is the number of data points in dimension $D$, $K$ is the cluster number, and $I$ is the number of iterations. 

Another popular method that builds on K-means is \textit{spectral clustering}.  Here, one leverages spectral graph theory to perform dimensionality reduction before applying the K-means algorithm. In spectral clustering, an undirected weighted graph $\mathcal{G}$ is formed, where each vertex is a data point and the edge weight is a chosen affinity between vertices. Let $W = (w_{i,j})_{i,j=1}^n$ denote the weighted adjacency matrix for the graph $\mathcal{G}$.  The \textit{(unnormalized) graph Laplacian} $L$ of $W$ is a matrix that captures the combinatorial properties of the Laplacian on discrete data.  The Laplacian $L$ is a symmetric positive semi-definite matrix, so the eigenvalues may be ordered $0 = \lambda_1 \leq \lambda_2 \leq \dots \leq \lambda_n$. Finding the eigenvectors $e_j$ corresponding to the lowest $K$ eigenvalues, define $U = [e_1| e_2 | \dots | e_K]$ and cluster the rows using K-means.  For more details, see \cite{ng2002spectral}. 

A bottleneck of spectral clustering is its complexity.  Given $N$ data points, forming the adjacency matrix and computing eigenvectors carries a computational complexity of $\mathcal{O}(N^3)$. Given K-means' cost is $\mathcal{O}(NDKI)$, the total complexity of spectral clustering is $\mathcal{O}(N^3) + \mathcal{O}(NDKI)$.  Though the computational complexity is roughly $N^3$, our application of spectral clustering will be on a smaller dataset.

Beyond the clustering challenges discussed in the introduction, K-means (and therefore spectral clustering) require the user to choose the number of clusters $K$ in advance.  In most applications, this parameter isn't known, and additional heuristics are required to select this value. In spectral clustering, one can use the eigenvalues of the Laplacian $L$ to determine the cluster number.  Specifically, as the eigenvalues are ordered, search for a value of $K$ such that the first $\lambda_1, \dots \lambda_k$ are small, and $\lambda_{k+1}$ is large. This method is justified by the fact that the spectral properties of $L$ are closely related to the connected components of $\mathcal{G}$ \cite{von2007tutorial}.  
Use of graph Laplacian eigenvalues to decide the cluster number $K$ is called the \textit{eigen-gap} heuristic.

\subsubsection{Graph cut clustering}


Given a notion of distance of data, the adjacency graph or matrix records the pairwise similarity.   Clustering the data $X$ into $K$ clusters is equivalent to providing a $K-$cut of the adjacency graph $\mathcal{G}$.  Graph cut strategies vary depending upon application.  For example, the min cut algorithm minimizes the cost between components of the graph, but this can result in an undesirable clustering, e.g., a cluster with one element. 

The \textit{Ratio cut} is a graph cut that seeks to ameliorate this issue by incorporating the size of each component.  Concretely, let $I \subset \{1,2, \dots , n\}$, $I^c$ is the complement of $I$, and let $W(I):=\sum_{i \in I, j \in I^c} w_{i,j}$ be the total cost between the component $I$ and its complement.  Given disjoint subsets $I_1, I_2, \dots I_k$ such that $\bigcup I_j = \{1, 2, \dots , n\}$, its \textit{ratio cut} is defined as
\begin{equation}
    \label{ratio}
  \mbox{RC}(I_1, I_2, \dots, I_k):= \frac{1}{2} \sum_{i=j}^k \frac{W(I_j)}{|I_j|}.
\end{equation}

Finding $I_1, \dots I_k$ such that Equation \ref{ratio} is minimized is NP-hard \cite{wagner1993between}.  Luckily, the optimization problem in Equation \ref{ratio} can be written as a trace minimization problem involving a particular matrix arising from the graph cut. By relaxing this optimization problem to one where we don't fix the form of the matrix in the trace, one obtains an optimization problem whose solution can be obtained by spectral clustering.  See \cite{von2007tutorial} for further details.

\subsection{Usage of K-means and Ratio-Cut in This Work}

In the subsequent implementation of CGC and MIER algorithms in Section 4, we will require both a clustering algorithm for CGC and a graph cut algorithm for MIER.  In our implementation of CGC, we will use K-means algorithm (executed with Llyods algorithm) for clustering, and in MIER we will use Ratio-Cut (executed by spectral clustering). We have made these choices because:

\begin{enumerate}
    \item K-means has historical precedence in clustering for climate applications \cite{netzel2016using,zscheischler2012climate} and straightforward implementation.  The scope of this work is not to find the ``best clustering'' of our data;  instead, we wish to understand how coarse-graining effects clustering and can be used to develop an ensemble of clusterings for use in understanding cluster method sensitivity to latent data scales.
    \item The MIER algorithm will require a graph cut of a particular adjacency matrix formed by a large ensemble of coarse-grain clusterings.  As discussed in Section 3.2, the adjacency graph is formed using normalized mutual information (see Section 2.3 for Mutual Information).  The ratio cut creates reasonable large components of this graph.  Each component will be heterogenous, and distinct from one another on an information-theoretic level. 
\end{enumerate}

\subsection{Mutual Information}


 Mutual information provides a method to quantify the shared information. Here, we outline how the mutual information is computed.  For a more detailed account of mutual information and clustering, see \cite{dom2002information} and \cite{vinh2010information}.

Let $X = \{x_i\}_{i=1}^n$ be a collection of data points. Suppose that we partition the data $X$ into two clusterings $U = \{U_i\}_{i=1}^k$ and $V = \{V_i\}_{i=1}^l$.  The \textit{entropy} of the clustering $U$, denoted $\mathcal{H}(U)$ is the average amount of information (e.g., in bits) needed to encode the cluster label for each data points of $U$.  If the clustering $V$ is known,  $U$ can be encoded with less bits of information.   The \textit{conditional entropy} $\mathcal{H}(U|V)$ denotes the average amount of information needed to encode $U$ if $V$ is known.

The \textit{mutual information} $\mathcal{I}(U,V)$ measures how knowledge of one clustering reduces our uncertainty of the other.  Formally, 
\[
\mathcal{I}(U,V) = \mathcal{H}(U) - \mathcal{H}(U|V).
\]
Explicit formulas for $\mathcal{H}(U)$ and $\mathcal{H}(U|V)$ can be derived as follows.  Let $n_{i,j} = |U_i \cap V_j|$ denote the number of points in both $U_i$ and $V_j$.  We set $a_i  = |U_i| = \sum_{j=1}^l n_{i,j}$ to be the size of $U_i$, and $b_j = |V_j| =  \sum_{i=1}^k n_{i,j}$ to be the size of $V_j$. 

Assume points of $X$ are sampled uniformly. Then the probability that a random point in $x \in X$ is in cluster $U_i$ is $p(x) = \frac{a_i}{n}$. Moreover, the probability that points $x,y \in X$ satisfy $x \in U_i$ and $y \in V_j$ is $p(x,y) = \frac{n_{i,j}}{n}$. Therefore, it follows that 
\begin{gather*}
    \mathcal{H}(U)  =  -\sum_{x \in U} p(x) \log(p(x))  =  - \sum_{i=1}^k \frac{a_i}{n} \log \left( \frac{a_i}{n} \right),\\
    \\
\mathcal{H}(U|V)  = - \sum_{x \in U, y \in V} p(x,y) \log \left( \frac{p(x,y)}{p(y)} \right)   =  - \sum_{i=1}^k \sum_{j=1}^l \frac{n_{i,j}}{n} \log \left( \frac{n_{i,j}/n}{b_j/n} \right),
\end{gather*}
which yields, 
\[
\mathcal{I}(U,V) = \sum_{i=1}^k \sum_{j=1}^l \frac{n_{i,j}}{n} \log \left( \frac{n_{i,j}/n}{a_i b_j/n^2} \right).
\]
As the values of $n_{i,j}$, and therefore $a_i,b_j$ are determined by the cluster values for the $N^2$ pairs of datapoints, the complexity to compute the mutual information is $\mathcal{O}(N^2)$.  Notice that $\mathcal{I}(U,V) \geq 0$, and $\mathcal{I}(U,V) = \mathcal{I}(V,U)$.  It then follows that
\begin{equation}
\label{normalize}
\mathcal{I}(U,V) \leq \min(\mathcal{H}(U), \mathcal{H}(V))) \leq \frac{1}{2} (\mathcal{H}(U) + \mathcal{H}(V) ).
\end{equation}
Equation \ref{normalize} shows that we can normalize the mutual information by dividing by the average of the entropies \cite{vinh2010information}.  Throughout, we define the \textit{normalized mutual information} as
\[
\mathcal{NI}(U,V) := \frac{2 \mathcal{I}(U,V)}{\mathcal{H}(U) + \mathcal{H}(V)}.
\]

We will use the (normalized) mutual information as a way to measure how similar two clusterings are. Since mutual information quantifies how much information from one clustering is obtained from another, it is ideal for measuring how much structure is lost by coarse-graining. As such, mutual information will play a key role in the MIER algorithm for ensemble selection. A $\mathcal{NI}(U,V)$ value of $p$ can be interpreted as $100\times p$ percent of information was ``lost'' between the clusters $U$ and $V$. We will be using this interpretation in Section 4. 

\section{The CGC and MIER Algorithms}



Here, we present our wavelet-based clustering model for classifying slices of tensor data. We detail the clustering algorithm  \textit{Coarse-Grain Clustering} (CGC) and present a method for selecting clusters to include in an ensemble based off the mutual information, which we call \textit{Mutual Information Ensemble Reduce} (MIER).  Table \ref{notation} contains all the notation used throughout both algorithms for ease of reference.

\begin{center}
\begin{tabular}{|c|c|}
\hline
Notation - CGC & Description\\
\hline
$\tX = \left( \tX_{i_1, i_2, i_3, i_4} \right)_{i_1, i_2, i_3, i_4 = 1}^{N_1, N_2, N_3, N_4}$ & Tensor of climate data\\
$\tX_{l}$, $l=1, \dots N_4$ & Tensors via fixing $i_4$ index\\
$\Psi_j$, $j=1,2,3$ & Wavelet for the indices $i_1, i_2, i_3$\\
$\ell_j$, $j=1,2,3$  & Scale level of the DWT on index $i_j$\\
\hline
Notation - MIER & Description\\
\hline
$\mathcal{L} \subset \mathbb{N}^3$ &  Permissible set of wavelet resolutions $(\ell_1, \ell_2, \ell_3)$\\
$\vec{\ell}$ & A permissible point $(\ell_j)_{j=1}^3 \in \mathcal{L}$\\
$U^{\vec{\ell}}$ & CGC at resolution $\vec{\ell}$\\
$\mathcal{G}$ & $\mathcal{NI}$ graph for $\{U^{\vec{\ell}} \}_{\ell \in \mathcal{L}}$ \\
$W$ & Weighted adjacency matrix for $\mathcal{G}$\\
$\{\mathcal{L}_j\}_{j=1}^K$ & Components of $\mathcal{L}$ corresponding to $K-$cut of $\mathcal{G}$\\
$\mathcal{A}(U^{\vec{\ell}})$ &  Average $\mathcal{NI}$ between $U^{\vec{\ell}}$ and its component \\
$\{U^j\}_{j=1}^K$ & Reduced ensemble - output of MIER \\
\hline
\hline
\end{tabular}
\captionof{table}{Notation for CGC and MIER Algorithms} 
\label{notation} 
\end{center}

\subsection{Coarse-Grain Clustering (CGC)}


This manuscript considers 4-way climate data tensors $\tX \in \R^{N_1 \times N_2 \times N_3 \times N_4}$.  We will index the modes of the tensor  using subscripts, namely
\[
\tX = \left( \tX_{i_1, i_2, i_3, i_4} \right)_{i_1, i_2, i_3, i_4 = 1}^{N_1, N_2, N_3, N_4}.
\]
Each of the coordinates coordinates $i_1, \dots, i_4$ describes a feature of the abstract dataset $\tX$.  Correspondingly, we will always make the following physical identifications:  the first and second indices $i_1$ and $i_2$ refer to latitude and longitude coordinates, respectively; the index $i_3$ denotes time, and $i_4$ refers to state variables (e.g., temperature or precipitation).

The goal of this work is to provide meaningful clusterings for the spatial location, namely the coordinates corresponding to $i_1$ and $i_2$.  
Hence, we seek clusterings of the indices $(i_1, i_2) \in \{1, 2, \dots, N_1 \} \times \{1, 2, \dots , N_2 \}$ using the data $\tX$.   While our focus is on clustering two indices of 4-way tensors, we note that this method does generalize to clustering d-way tensors along any number of indices. 

\begin{figure}[h!]
    \centering
    \includegraphics[width = \textwidth]{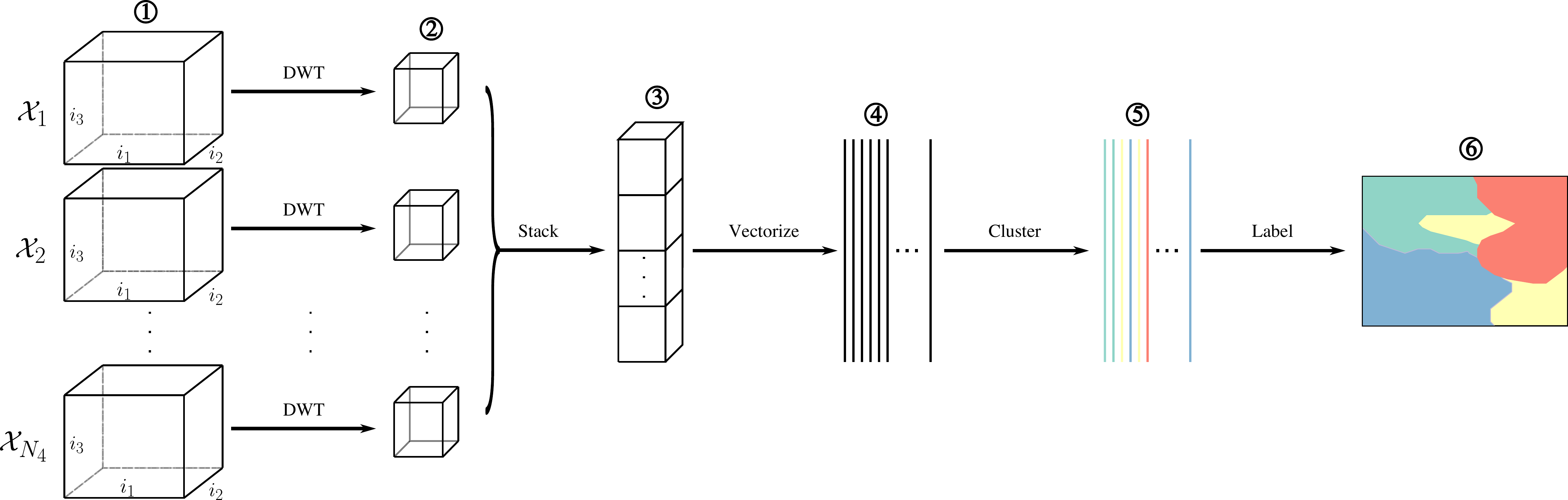}
    \caption{Diagram of each step of the CGC algorithm.  }
    \label{fig:mr_cluster}
\end{figure}

 We now describe the Coarse-Grain Clustering algorithm. Figure \ref{fig:mr_cluster} schematically displays the key features of CGC, while Algorithm \ref{alg:mr_cluster} contains the pesudo-code.

\textbf{Step One - Split Tensor:} The first step in the coarse-grain clustering (CGC) algorithm is to separate the tensor $\tX$ into  sub-tensors that are largely statistically uncorrelated across the dataset. For example, temperature and precipitation are locally correlated - e.g., seasonal rainfall.  However, they are weakly correlated at large spatial scales. Indeed, there are hot dry deserts, cold dry deserts, wet cold regions, and wet hot regions. Therefore in the climate dataset $\tX$, one would separate by climate variables, but not by space or time.  In a generic, non-climate specific tensor,  one might split across different variables or runs of an experiment.    We let $\tX_1, \tX_2, \dots , \tX_{N_4}$ be the 3-way tensors obtained by fixing the $i_4$ index to the $N_4$ possible values. Note that each of these tensors $\tX_{l}$ for $l = 1, \dots, N_4$ have the same size. 

\textbf{Step Two - DWT:} After splitting the tensor $\tX$ into sub-tensors, the next step is to select the inputs. The user chooses wavelets for each of the remaining indices $i_1, i_2$ and $i_3$. We let $\Psi_j$ denote the wavelet for the index $i_j$, $j=1,2,3$.  Non-negative integers $\ell_j$ for $j=1,2,3$ are selected to control the scale level of the DWT on index $i_j$. For each 3-way climate variable tensor $\tX_{l}$, $l=1, \dots N_4$,  take the DWT transform.  

\textbf{Step Three - Stack:} Since the same wavelets are used on each $\tX_{l}$, the DWT of $\tX_{l}$ will each have the same shape.  These tensors can therefore be stacked along the face we wish to classify.  For the climate biome problem, this would be the $(i_1, i_2)$ face. 

\textbf{Step Four - Vectorize:} Once the approximation coefficients are stacked, they may be vectorized along the face of interest.  These vectors will be clustered according to a clustering algorithm of choice. This will result in a clustering of the face of interest on the DWT stack.  

\textbf{Step Five - Clustering:} The final input is the choice of clustering algorithm, as well as any hyper-parameters required for the chosen algorithm. For example, one may choose K-means, in which case the user needs to specify the number of clusters $k$. Let $\mathcal{C}$ denote the chosen clustering algorithm, along with its chosen hyper-parameters. With the inputs chosen, the algorithm proceeds as follows.  Algorithm $\mathcal{C}$ is applied to the vectorized DWT coefficients from step four.

\textbf{Step Six - Return Labels:}  The final step is to translate these labels on the coarse-grain stack to the face of the original data set. This is done using the inverse DWT. Specifically, cluster labels corresponding with the largest value appearing in the inverse DWT filter are used to propagate the coarse label to finer detail.

\begin{algorithm}[H]
\label{alg:mr_cluster}
 \KwIn{$\tX =  \left( \tX_{i_1, i_2, i_3, i_4} \right)_{i_1, i_2, i_3, i_4 = 1}^{N_1, N_2, N_3, N_4}$, $\{\Psi_j, \ell_j\}_{j=1}^3$,  $\mathcal{C}$}
 \KwResult{Clustering of $(i_1, i_2)$ face}
 form tensors $\tX_l$, $l=1, \dots N_4$ \;
 take DWT of each $\tX_l$\;
 stack approximation coefficents along $(i_1, i_2)$ face\;
 vectorize stack of DWT\;
 cluster vectors according to algorithm $\mathcal{C}$\;
 return labels to $(i_1, i_2)$ face
 \caption{Coarse-Grain Cluster}
\end{algorithm}

The complexity of CGC  determined by the DWT and the clustering algorithm $\mathcal{C}$.  The cost of the DWT (and its inverse) is $\mathcal{O}(N_1 N_2 N_3)$.  The cost of $\mathcal{C}$ depends, of course, on the choice of clustering algorithm. For example, suppose we use K-means (as we will in our applications).  Then since the size of the approximation coefficients is $\hat{N} := \frac{N_1}{2^{\ell_1}}\frac{N_2}{2^{\ell_2}}\frac{N_3}{2^{\ell_3}}$, the total cost of CGC is $\mathcal{O}(N_1 N_2 N_3) + \mathcal{O}(K \hat{N} I)$.  

We remark that the process of computing coarse-grain clusterings is extremely parallelizable. Indeed, a directed tree structure can be implemented to semi-parallelize all scales of the DWT.  Each of these scales can then be independently clustered. This fact, combined with the extreme size savings of the DWT, means the cost to compute many CGC's is no worse than computing the most expensive CGC in the collection.

\subsection{Mutual Information Ensemble Reduce (MIER)}

The CGC algorithm describes how to produce a single clustering at a fixed coarse-graining. This coarse-graining arises from the choice of wavelets and wavelet levels $\{\Psi_j, \ell_j\}_{j=1}^3$. {\bf The power behind CGC is its ability to produce many clusterings by simply varying the wavelet levels $\ell_j$, $j=1,2,3$}, which as discussed above, can be readily parallelized via a single instruction.

This process results in a large ensemble of clusters, one that is potentially too big to analyze. In this section, we discuss a method to select a small subset of this large ensemble of coarse-grained clusters. Our method  leverages the mutual information to find a compact subset of clusters that contains most of the information across the large ensemble.  This is accomplished by computing the mutual information between all the clusters in the large ensemble. This results in a connected graph.  This connected graph is then ratio-cut to find heavily connected and therefore information theoretically similar clusters. For each component, we again use mutual information to select a single representative of the component.  We call this method \textit{Mutual Information Ensemble Reduce} (MIER).

The MIER algorithm is summarized in Figure \ref{fig:en_red} and Algorithm \ref{alg:en_red}.  The details of the algorithm are as follows.

\begin{center}
\centering
\begin{figure}[h!]
\centerfloat
    \includegraphics[width=\textwidth]{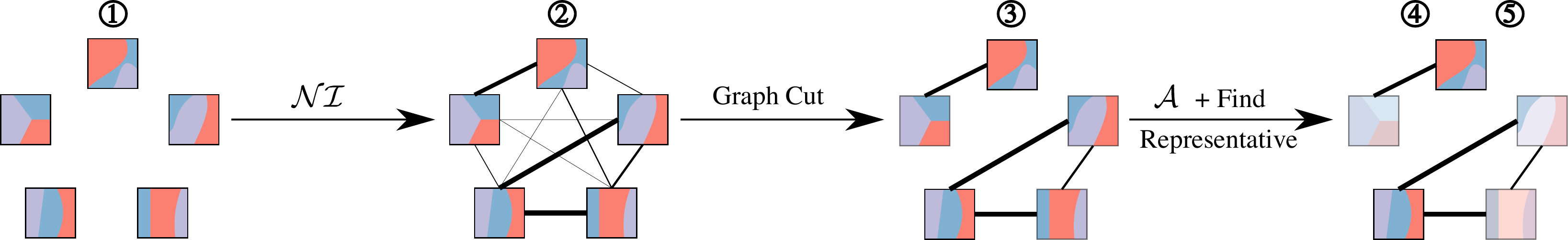}
    \caption{Diagram of each step of the MIER algorithm. In step (2), thicker lines correspond to larger mutual information. In step (5), the clustering with largest $\mathcal{A}(U{^{\vec{\ell}}})$ is highlighted. }
    \label{fig:en_red}
\end{figure}
\end{center}

\textbf{Step One - Large Ensemble:} Let $\mathcal{L} \subset \mathbb{N}^3$ denote the permissible set of wavelet resolutions $(\ell_1, \ell_2, \ell_3)$ for the chosen wavelets $\{\Psi_j, \ell_j\}_{j=1}^3$.  Reasonable values for $\mathcal{L}$ can be deduced from the dataset and problem of interest, e.g. scale of data and anticipated importance of embedded features.   Once $\mathcal{L}$ has been decided, CGC is run for each $\vec{\ell}= (\ell_j)_{j=1}^3 \in \mathcal{L}$. We denote the clustering using the wavelet resolutions by $U^{\vec{\ell}}$.  This results in an ensemble of clusters $\{U^{\vec{\ell}}\}_{\vec{\ell} \in \mathcal{L}}$.  

\textbf{Step Two - Mutual Information:} Next, we compute the normalized mutual information between each clustering $ U^{\vec{\ell}}$ in our ensemble $\mathcal{L}$. This results in a complete weighted graph $\mathcal{G}$ on nodes indexed by the set $\mathcal{L}$.  The weight between node $\vec{\ell} = (\ell_1, \ell_2, \ell_3)$ and node $\vec{\ell'} = (\ell_1', \ell_2', \ell_3')$ is the normalized mutual information $\mathcal{NI}(U^{\vec{\ell}}, U^{\vec{\ell'}})$.  We call the graph $\mathcal{G}$ the \textit{mutual information graph}, and let $W$ denote the weighted adjacency matrix for $\mathcal{G}$.

\textbf{Step Three - Graph Cut:} Having built the mutual information graph, we now perform a graph cut.  Recall, 
spectral clustering solves a relaxed version of the ratio cut problem. Hence, we use spectral clustering on $W$ to find a ratio-cut of $\mathcal{G}$. The eigen-gap heuristic is used when selecting the number of clusters $K$ for spectral clustering $W$ \cite{von2007tutorial}.  Let $\mathcal{L}_1, \mathcal{L}_2, \dots \mathcal{L}_K$ denote the $K$ components of $\mathcal{L}$ corresponding to the $K-$cut of $\mathcal{G}$.

\textbf{Step Four - Average $\mathcal{NI}$:} For each component of the cut mutual information graph, we seek a best representative. Let $\mathcal{A}(U^{\vec{\ell}})$ denote the average mutual information between $U^{\vec{\ell}}$ and all other members of its component. That is, for  $\vec{\ell} \in \mathcal{L}_j$,
\[
\mathcal{A}(U^{\vec{\ell}}) = \frac{1}{|\mathcal{L}_j| - 1} \sum_{\vec{\ell'} \in \mathcal{L}_j, \vec{\ell'} \neq \vec{\ell}} \mathcal{NI}(U^{\vec{\ell}}, U^{\vec{\ell'}})
\]
where $\mathcal{NI}(U^{\vec{\ell}}, U^{\vec{\ell'}})$ is the normalized mutual information between the clusters $U^{\vec{\ell}}$ and  $U^{\vec{\ell'}})$.

\textbf{Step Five - Choose Representative:} For each $j=1, \dots K$, the goal is to select the clustering $U^{\vec{\ell}}$ that best represents all the clusterings in $\mathcal{L}_j$. If $U^{\vec{\ell}}$ is a good representative for all the other clusterings within its component, then the mutual information between $U^{\vec{\ell}}$ and the other members of the component will be high on average. Thus, $\mathcal{A}(U^{\vec{\ell}})$ will be large. Consequently, we select a cluster in $\mathcal{L}_j$ for which $\mathcal{A}$ is maximized:
\[
U^j \in \mbox{Argmax}_{\vec{\ell} \in \mathcal{L}_j} \mathcal{A}(U^{\vec{\ell}}).
\]
In the unlikely event that the Argmax consists of more than one element, one is selected randomly.

\begin{algorithm}[H]
\label{alg:en_red}
 \KwIn{$\{U^{\vec{\ell}}\}_{\vec{\ell} \in \mathcal{L}}$}
 \KwResult{$\{U^j \}_{j=1}^k$}
 import the large cluster ensemble $\{U^{\vec{\ell}}\}_{\vec{\ell} \in \mathcal{L}}$\;
 build mutual information graph $\mathcal{G}$\;
ratio cut $\mathcal{G}$\;
Compute $\mathcal{A}(U^{\vec{\ell}})$ for each $\vec{\ell} \in \mathcal{L}$\;
 return reduced ensemble $\{U^j \}_{j=1}^k$
\caption{Mutual Information Ensemble Reduce}
\end{algorithm}

To compute the complexity of  the MIER algorithm, we note there are $|\mathcal{L}|^2$ pairs of clusterings whose normalized mutual information needs to be computed.  If there are $N$ total data-points, the cost to form $\mathcal{G}$ is $\mathcal{O}(|\mathcal{L}|^2 N^2)$.  Next, spectral clustering on the $|\mathcal{L}|^2$ connections of the graph needs to be performed.  This costs $\mathcal{O}(|\mathcal{L}|^3) + \mathcal{O}(|\mathcal{L}|KI)$. This puts the total cost of MIER at  $\mathcal{O}(|\mathcal{L}|^2 N^2)+ \mathcal{O}(|\mathcal{L}|^3) + \mathcal{O}(|\mathcal{L}|KI)$. 

This cost is deceptively high however. In general, $|\mathcal{L}|$ is very small compared to $N$.  Indeed, $|\mathcal{L}|$ might be on the order of 10's or worst case 100's, while $N$ is many orders of magnitude higher. The complexity of the MIER algorithm is therefore dominated by the cost normalized mutual information, and is really more like $\mathcal{O}(N^2)$. 

\section{Application - Gridded Climate Dataset}


As a proof of concept, we apply the MR-Cluster to a gridded historical climate data set of North America \cite{livneh2015spatially}, referred to hereafter as L15.  This data set ingests station data and interpolates results for each grid point, integrating the effects of topography on local weather patterns. The gridded data is six by six kilometers a side and consists of 614 latitudinal, 928 longitudinal, and 768 temporal steps for the years 1950-2013. The available monthly variables in the L15 data set are averaged values of daily total precipitation, daily maximum temperature, daily minimum temperature, and daily average wind speed. A representative snapshot of precipitation, maximum and minimum temperature is shown in Figure \ref{fig:L15}. The datasets contains key inputs needed for biome classification using the KG model \cite{kottek2006world} and allows ready comparison against this expert judgement based approach. As this dataset is freely available, as well as widely used within the climate community (e.g., Henn et al. 2017), it provides a good benchmark application to illustrate capabilities of the method.

\subsection{CGC Hyperparameter Selection for L15}


The first step of CGC is to split the tensor $\tX$ into sub-tensors corresponding to the climate variables. The historical precedent has been to use temperature and precipitation data to prescribe the biomes \cite{kottek2006world, netzel2016using, thornthwaite1948approach, zscheischler2012climate}.  Hence, we will only consider the sub-tensors $\tX_1, \tX_2, \tX_3$ corresponding to precipitation and temperature values in the data set - namely the averaged values of daily total precipitation, daily maximum temperature, daily minimum temperature for each month. The next step is to determine the inputs to Algorithms \ref{alg:mr_cluster} and \ref{alg:en_red}.  We describe these now.

L15 is a gridded observational dataset that achieves a six km spatial resolution, while each time slice of the data represents monthly timescale data. 
Whenever a wavelet transform is taken, the spatial and/or temporal scales are approximately doubled. 
For example, the L15 dataset has a six  km spatial resolution.  Thus, the coarse wavelet coefficients have a spatial resolution of $12, 24, 48$, etc., km for one, two, and three wavelet transforms respectively. Similarly, wavelet transforms of the monthly time scales will result in $2,4,8$, etc., month long scales. 

\begin{center}
\centering
\begin{figure}[h!]
\centerfloat
    \begin{tabular}{ccc}
     \subfloat[Precipitation (mm)]{\includegraphics[width = 0.3\textwidth]{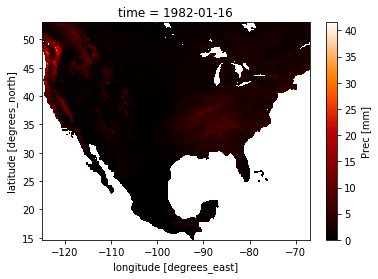}} &  \subfloat[Max Temp. ($^\circ C$)]{\includegraphics[width = 0.3\textwidth]{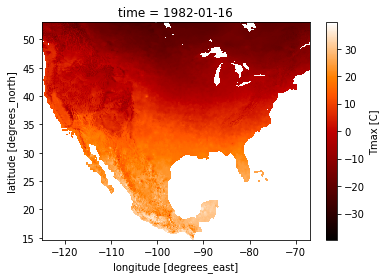}} &  \subfloat[Min Temp. ($^\circ C$)]{\includegraphics[width = 0.3\textwidth]{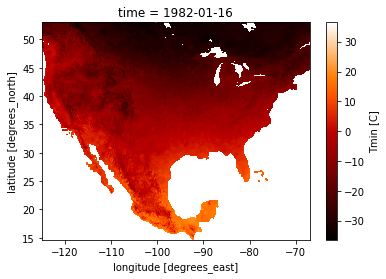}}
    \end{tabular}
    \caption{Representative variables within the from L15 dataset, illustrating broad range of both coarse and fine spatial scales for precipitation, maximum temperature, and minimum temperature.}
    \label{fig:L15}
\end{figure}
\end{center}

There is a scale at which both the spatial and temporal information is too coarse and begins to lose meaning.  For example, on one extreme the spatial scale of the entire dataset is meaningless.  On the other, the six km initial resolution is too fine scale for adequate characterizations into distinctly visible biomes at the North American scale.

These scales demarcate our set of permissible wavelet resolutions $\mathcal{L}$. At least one wavelet transform is taken in both space and time.  The maximum for the spatial indices $\ell_1, \ell_2$ is  four (roughly $96$ km).  The maximum number of temporal wavelet transforms is six (roughly $5$ years). 
Further, we opted for a parsimony with regards to the spatial wavelet transforms--  a wavelet transform is taken along $i_1$ (latitude) if and only if it is also taken along $i_2$ (longitude). For example, if we take two wavelet transforms in space laterally, we will also take two in space longitudinally so that horizontal spatial resolution is uniformly scaled.
Thus, we have 
\begin{equation}
    \label{lattice}
    \mathcal{L} = \{ (i,i,j): i=1, \dots , 4, j=1, \dots, 6\}.
\end{equation}

Note, while it was possible to push the maximum levels to coarser grain, we wanted to avoid the risk of over-coarsening the result. For our choice of wavelets, we choose Daubechies 2 (db2)  to match the time signals and Haar for space, corresponding to anticipated smooth periodicities in time and sharp gradients, e.g., near mountains, in space. 

For the algorithm $\mathcal{C}$, we have chosen to use K-means clustering for $k=4,5,\dots20$ due to the historical precedence this algorithm has in clustering for climate applications \cite{netzel2016using,zscheischler2012climate} and straightforward implementation.  Recall that the aim is not to find the ``best clustering'' of our data;  instead, we wish to understand how coarse-graining effects clustering and can be used to develop an ensemble of clusterings for use in understanding cluster method sensitivity to latent data scales.   


\subsection{Results}

\subsubsection{L15 CGC algorithm results - effects of coarse-graining.}


We begin with a qualitative analysis by visualizing  the effect scale has on clustering.  For a parsimonious exposition, we have opted to display the sensitivity of CGC on L15 to scale only for a fixed value of $K=10$. Different values of $K$ result in similar qualitative results.  Figure \ref{resolutions} explores this sensitivity.

Visually, scale can be seen to have a drastic effect on the overall structure of the clusters. For example, decreased temporal scale increases resolution from two to three eastern US classifications and shown in Figure \ref{ER_plots}c versus \ref{ER_plots}d and \ref{ER_plots}f. Coarsened classifications are observed as a direct role of spatial scale, e.g. Figure \ref{ER_plots}f versus \ref{ER_plots}c to \ref{ER_plots}e. Cluster boundary shape is also effected by the wavelet resolution.  For example, a vertical boundary can be found in the middle of the United States across each classification.  However, the shape of that boundary depends on the resolution, e.g.  Figure \ref{ER_plots}d versus \ref{ER_plots}e. 

\begin{center}
\centering
\begin{figure}[hbt!]
\centerfloat
    \begin{tabular}{cc}
        \subfloat[$(\ell_1, \ell_2, \ell_3) = (1,1,1)$]{\includegraphics[width = 0.45\textwidth]{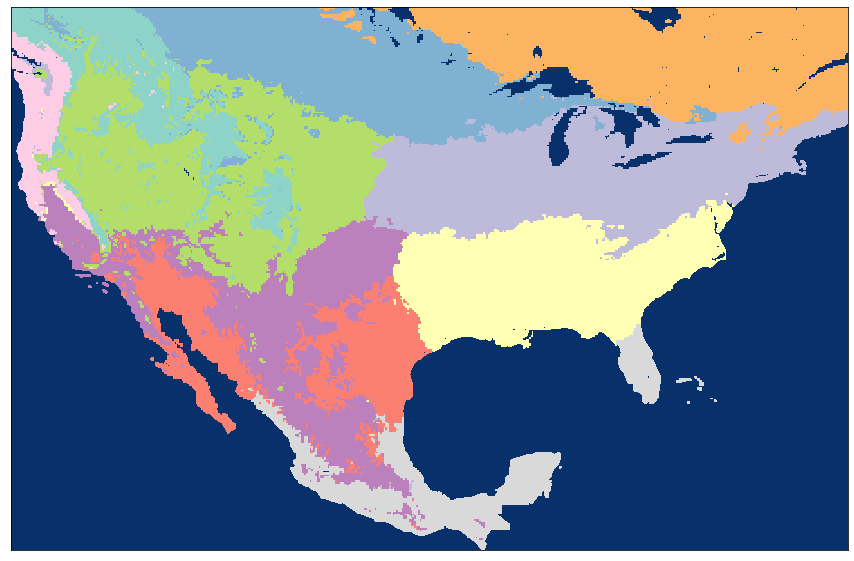}} &
        \subfloat[$(\ell_1, \ell_2, \ell_3)=  (4,4,1)$]{\includegraphics[width = 0.45\textwidth]{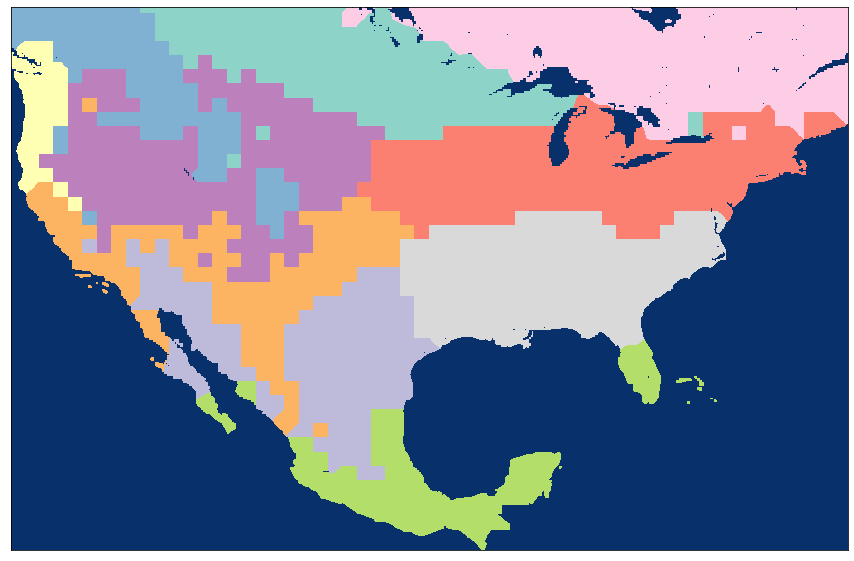}} \\

        \subfloat[$(\ell_1, \ell_2, \ell_3) = (1,1,3)$]{\includegraphics[width = 0.45\textwidth]{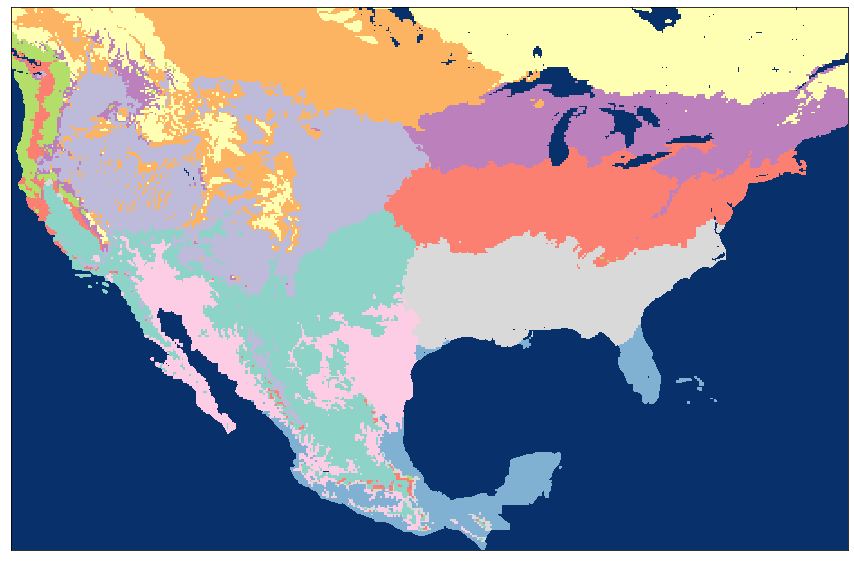}} &
        \subfloat[$(\ell_1, \ell_2, \ell_3) = (4,4,3)$]{\includegraphics[width = 0.45\textwidth]{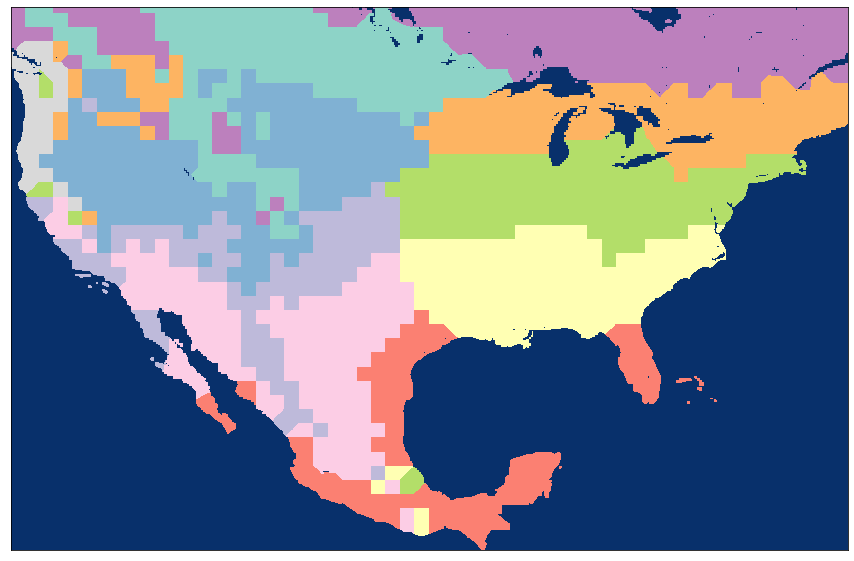}} \\

        \subfloat[$(\ell_1, \ell_2, \ell_3) = (1,1,6)$]{\includegraphics[width = 0.45\textwidth]{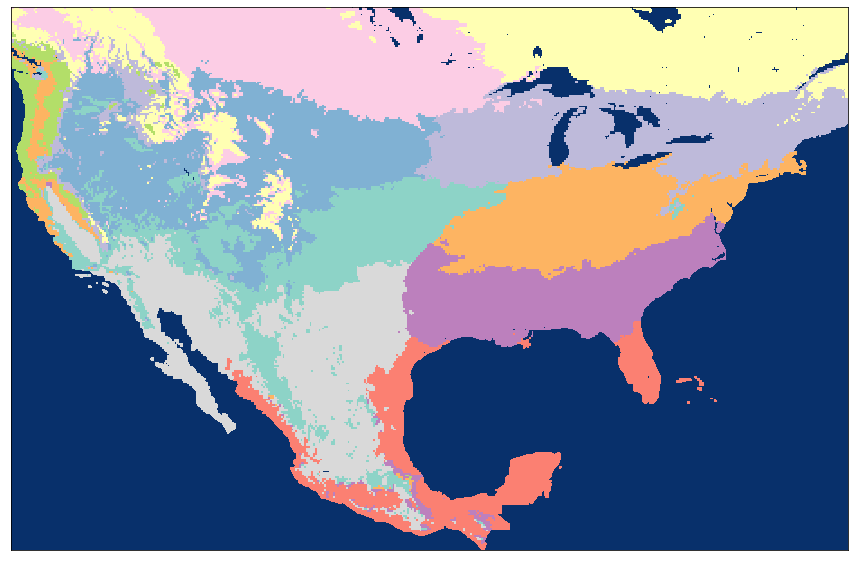}} &
        \subfloat[$(\ell_1, \ell_2, \ell_3) = (4,4,6)$]{\includegraphics[width = 0.45\textwidth]{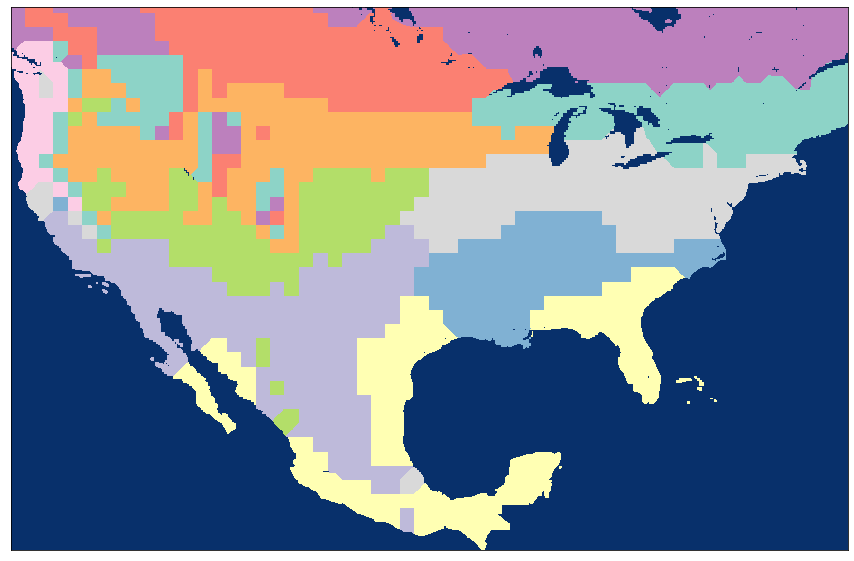}} \\
    \end{tabular}
    \caption{Clusterings obtained for $k=10$ at resolution $(\ell_1, \ell_2, \ell_3)$. }
    \label{resolutions}
\end{figure}
\end{center}

However, note that several coherent features are observed.  Strong latitudinal dependence in the eastern portion of the US is consistent across clusterings as scales are modified, e.g., Figure \ref{resolutions}a to \ref{resolutions}d.  Moreover, the location of the class boundary lines is relatively stable across all resolutions, in particular, in the Midwestern United States.

For a quantitative measure of the effects scale has on CGC for the L15 dataset, we use statistics derived from the normalized mutual information across various cluster numbers. Specifically, we compute the CGC clusterings $\{U^{\vec{\ell}}\}_{\ell \in \mathcal{L}}$ for each $K=4, \dots , 20$.  For each fixed $K$, we then computed the normalized mutual information between ``adjacent'' scales:
\[
\mathcal{NI}(U^{(1,1,1)}, U^{(1,1,2)}), \mathcal{NI}(U^{(1,1,1)}, U^{(2,2,1)}), \mathcal{NI}(U^{(1,1,2)}, U^{(1,1,3)}), \dots, \mathcal{NI}(U^{(3,3,6)}, U^{(4,4,6)}).
\]
By using adjacent resolutions, we are measuring the sensitivity of K-means to adjusting scale. Computing basic statistics across all $K$ allows us to infer the expected information loss in coarse-grain clustering. Table \ref{NMI_single} and Figures \ref{NMI_time}-\ref{NMI_PDF} present this statistical data at different levels of granularity.

In Table \ref{NMI_single}, we display $\mathcal{NI}$ pertaining to each possible adjacent coarse-graining. Since there are $38$ adjacent coarse-grainings, each with $16$ total $\mathcal{NI}$ values, a parsimonious representation is required.  For Table \ref{NMI_single}, we have chosen to only present the minimum, average, and maximum $\mathcal{NI}$ values.

Next, we begin grouping the data based on a fixed pair of adjacent spatial or temporal scale lengths.  
In Figure \ref{NMI_time}, we have binned all the $\mathcal{NI}$ values for adjacent spatial resolutions. Once the data is binned, the minimum, average, and maximum are computed, as well as a histogram and a kernel density estimate. For example, Figure \ref{NMI_time}a contains all the $\mathcal{NI}$ values between $(1,1,j)$ and $(2,2,j)$ for all $j=1, \dots , 6$ and all $K=4, \dots, 20$. The spread of these plots is the range of observed $\mathcal{NI}$. These are the worst and best case scenarios for information loss transitioning between the stated resolutions. The peaks correspond to values  $\mathcal{NI}$ that were frequently observed.  These are the approximate expected information loss transitioning between the stated resolutions. 

In Figure \ref{NMI_space}, we follow the same procedure only binning temporal scales instead of spatial scales. Finally, Figure \ref{NMI_PDF} takes the entire collection of $\mathcal{NI}$ values, plotting the PDF as well as the minimum, average and maximum.

\renewcommand{\arraystretch}{2}
\begin{table}[hbt!]
\resizebox{\textwidth}{!}{%
\begin{tabular}{ccccccc}
\cline{1-1} \cline{3-3} \cline{5-5} \cline{7-7}
\multicolumn{1}{|c|}{\textbf{(1,1,1)}} & \multicolumn{1}{c|}{\{0.83, 0.89, 0.91\}} & \multicolumn{1}{c|}{\textbf{(2,2,1)}} & \multicolumn{1}{c|}{\{0.77, 0.88, 0.94\}} & \multicolumn{1}{c|}{\textbf{(3,3,1)}} & \multicolumn{1}{c|}{\{0.74, 0.79, 0.83\}} & \multicolumn{1}{c|}{\textbf{(4,4,1)}} \\ \cline{1-1} \cline{3-3} \cline{5-5} \cline{7-7} 
\{0.77, 0.88, 0.94\}                   &                                           & \{0.77, 0.88, 0.94\}                  &                                           & \{0.78, 0.88, 0.95\}                  &                                           & \{0.72, 0.88, 0.97\}                  \\ \cline{1-1} \cline{3-3} \cline{5-5} \cline{7-7} 
\multicolumn{1}{|c|}{\textbf{(1,1,2)}} & \multicolumn{1}{c|}{\{0.83, 0.87, 0.89\}} & \multicolumn{1}{c|}{\textbf{(2,2,2)}} & \multicolumn{1}{l|}{\{0.78, 0.84, 0.87\}} & \multicolumn{1}{c|}{\textbf{(3,3,2)}} & \multicolumn{1}{c|}{\{0.66, 0.77, 0.83\}} & \multicolumn{1}{c|}{\textbf{(4,4,2)}} \\ \cline{1-1} \cline{3-3} \cline{5-5} \cline{7-7} 
\{0.7, 0.77, 0.87\}                    &                                           & \{0.73, 0.77, 0.86\}                  &                                           & \{0.71, 0.76, 0.87\}                  &                                           & \{0.64, 0.74, 0.87\}                  \\ \cline{1-1} \cline{3-3} \cline{5-5} \cline{7-7} 
\multicolumn{1}{|c|}{\textbf{(1,1,3)}} & \multicolumn{1}{c|}{\{0.73, 0.82, 0.87\}} & \multicolumn{1}{c|}{\textbf{(2,2,3)}} & \multicolumn{1}{c|}{\{0.76, 0.81, 0.85\}} & \multicolumn{1}{c|}{\textbf{(3,3,3)}} & \multicolumn{1}{c|}{\{0.64, 0.73, 0.77\}} & \multicolumn{1}{c|}{\textbf{(4,4,3)}} \\ \cline{1-1} \cline{3-3} \cline{5-5} \cline{7-7} 
\{0.77, 0.89, 0.95\}                   &                                           & \{0.84, 0.91, 0.95\}                  &                                           & \{0.78, 0.88, 0.94\}                  &                                           & \{0.77, 0.86, 0.96\}                  \\ \cline{1-1} \cline{3-3} \cline{5-5} \cline{7-7} 
\multicolumn{1}{|c|}{\textbf{(1,1,4)}} & \multicolumn{1}{c|}{\{0.73, 0.82, 0.87\}} & \multicolumn{1}{c|}{\textbf{(2,2,4)}} & \multicolumn{1}{c|}{\{0.76, 0.81, 0.86\}} & \multicolumn{1}{c|}{\textbf{(3,3,4)}} & \multicolumn{1}{c|}{\{0.64, 0.71, 0.76\}} & \multicolumn{1}{c|}{\textbf{(4,4,4)}} \\ \cline{1-1} \cline{3-3} \cline{5-5} \cline{7-7} 
\{0.81, 0.86, 0.92\}                   &                                           & \{0.76, 0.86, 0.91\}                  &                                           & \{0.79, 0.86, 0.91\}                  &                                           & \{0.76, 0.83, 0.92\}                  \\ \cline{1-1} \cline{3-3} \cline{5-5} \cline{7-7} 
\multicolumn{1}{|c|}{\textbf{(1,1,5)}} & \multicolumn{1}{c|}{\{0.81, 0.84, 0.88\}} & \multicolumn{1}{c|}{\textbf{(2,2,5)}} & \multicolumn{1}{c|}{\{0.72, 0.78, 0.84\}} & \multicolumn{1}{c|}{\textbf{(3,3,5)}} & \multicolumn{1}{c|}{\{0.63, 0.71, 0.78\}} & \multicolumn{1}{c|}{\textbf{(4,4,5)}} \\ \cline{1-1} \cline{3-3} \cline{5-5} \cline{7-7} 
\{0.78, 0.83, 0.9\}                    &                                           & \{0.77, 0.82, 0.87\}                  &                                           & \{0.76, 0.82, 0.87\}                  &                                           & \{0.74, 0.79, 0.85\}                  \\ \cline{1-1} \cline{3-3} \cline{5-5} \cline{7-7} 
\multicolumn{1}{|c|}{\textbf{(1,1,6)}} & \multicolumn{1}{c|}{\{0.78, 0.83, 0.89\}} & \multicolumn{1}{c|}{\textbf{(2,2,6)}} & \multicolumn{1}{c|}{\{0.75, 0.8, 0.85\}}  & \multicolumn{1}{c|}{\textbf{(3,3,6)}} & \multicolumn{1}{c|}{\{0.67, 0.71, 0.79\}} & \multicolumn{1}{c|}{\textbf{(4,4,6)}} \\ \cline{1-1} \cline{3-3} \cline{5-5} \cline{7-7} 
\end{tabular}
}
\caption{$\mathcal{NI}$ between adjacent clusterings in scale, averaged across all clusterings $K=4, \dots , 20$. Values $\textbf{(i,i,j)}$ indicate the scale for CGC.  The values $\{\mbox{min}, \mbox{average}, \mbox{max}\}$ between adjacent scales are arising from the $16$ normalized mutual information.} 
\label{NMI_single} 
\end{table}
\renewcommand{\arraystretch}{1}


\begin{center}
\centering
\begin{figure}[hbt!]
\centerfloat
    \begin{tabular}{ccc}
        \subfloat[Scale $1-2$: $\{0.73, 0.84, 0.91\}$]{\includegraphics[width = 0.32\textwidth]{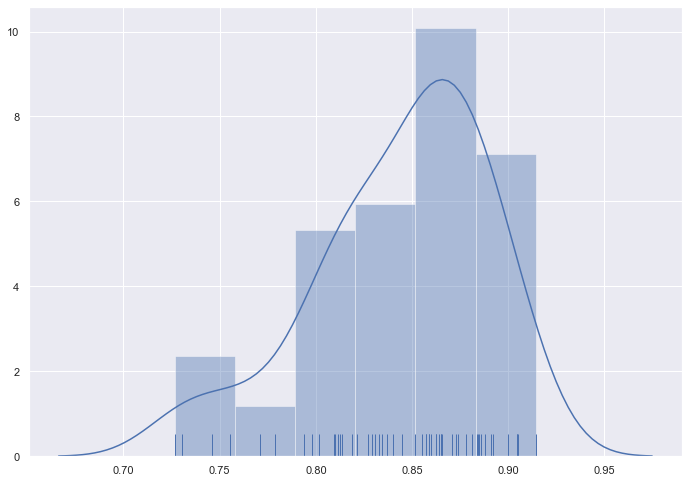}} &
        \subfloat[Scale $2-3$: $\{0.72, 0.82, 0.89\}$]{\includegraphics[width = 0.32\textwidth]{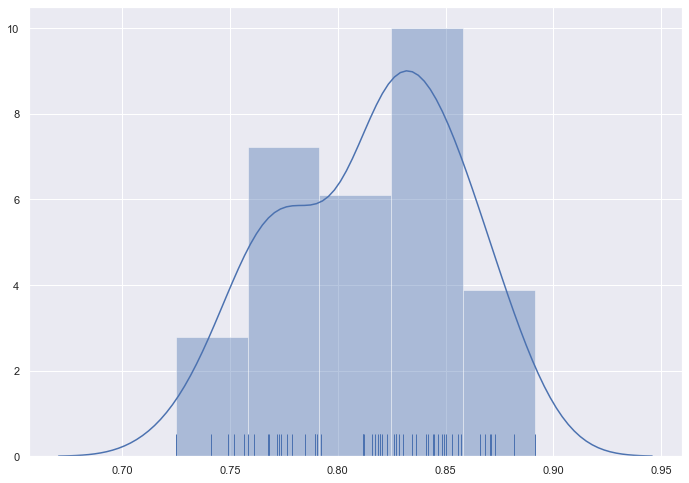}} &
        \subfloat[Scale $3-4$: $\{0.63, 0.74, 0.83\}$]{\includegraphics[width = 0.32\textwidth]{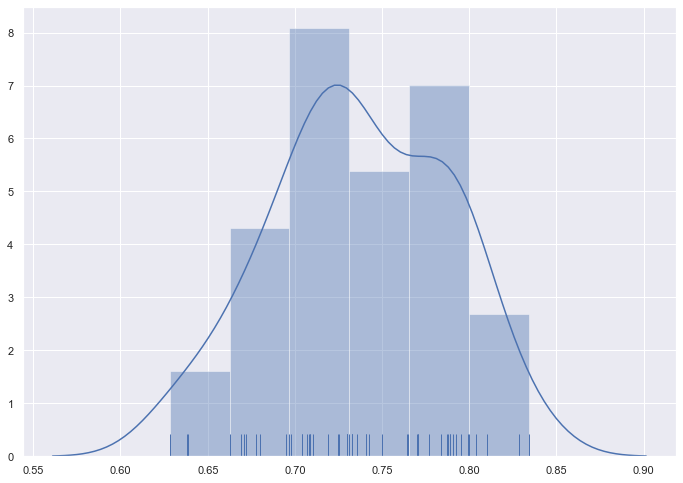}} 
    \end{tabular}
    \caption{$\mathcal{NI}$ values binned by spatial resolution. Scale $i-i+1$ indicates the levels of the spatial wavelet transform bin. The $\{\mbox{min}, \mbox{average}, \mbox{max}\}$ is displayed for each bin.}
    \label{NMI_time} 
\end{figure}

\end{center}

\begin{center}
\centering
\begin{figure}[hbt!]
\centerfloat
    \begin{tabular}{ccc}
        \subfloat[Scale $1-2$: $\{0.72, 0.88, 0.97\}$]{\includegraphics[width = 0.32\textwidth]{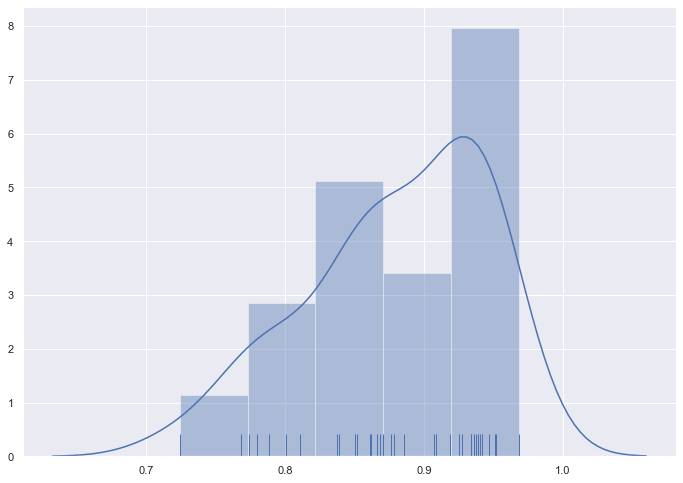}} &
        \subfloat[Scale $2-3$: $\{0.64, 0.76, 0.87\}$]{\includegraphics[width = 0.32\textwidth]{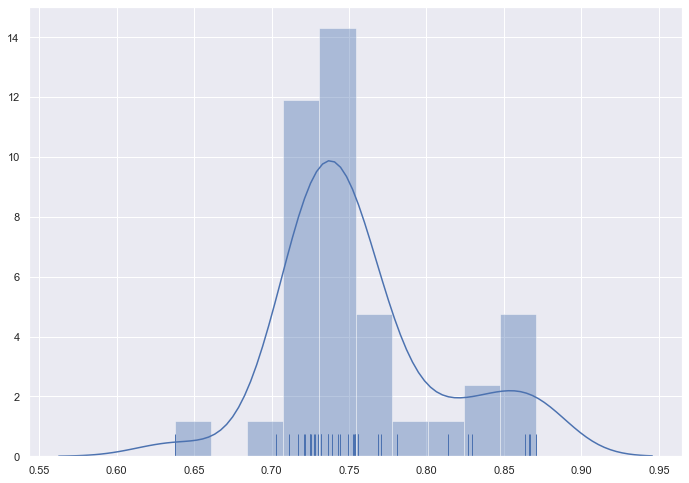}} &
        \subfloat[Scale $3-4$: $\{0.77, 0.88, 0.96\}$]{\includegraphics[width = 0.32\textwidth]{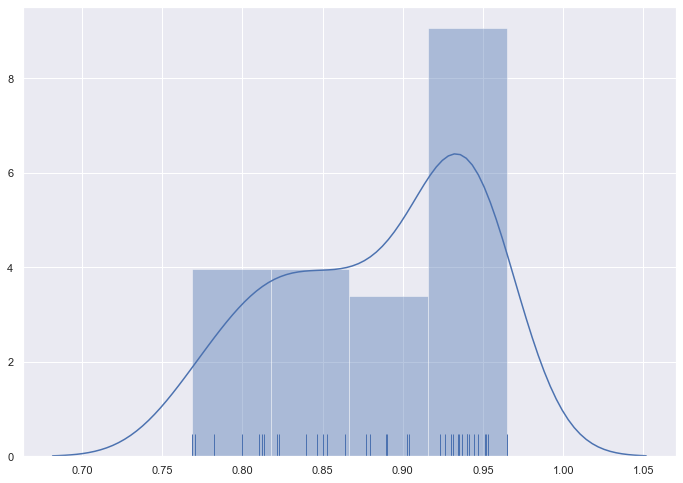}} \\
        \subfloat[Scale $4-5$: $\{0.76, 0.85, 0.92\}$]{\includegraphics[width = 0.32\textwidth]{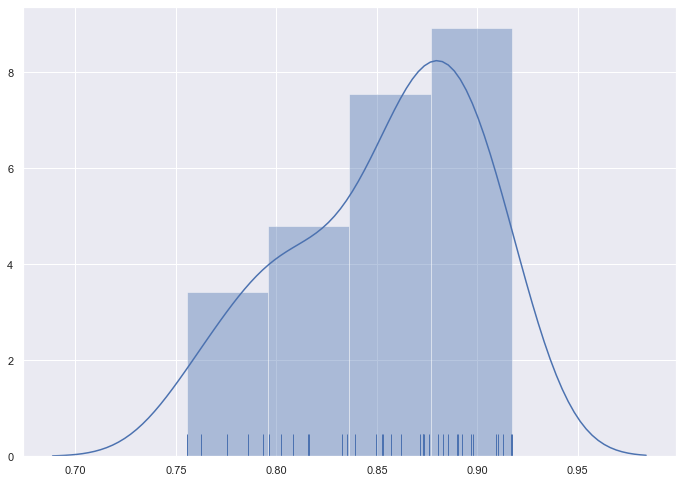}} & 
                     &
        \subfloat[Scale $5-6$: $\{0.74, 0.81, 0.9\}$]{\includegraphics[width = 0.32\textwidth]{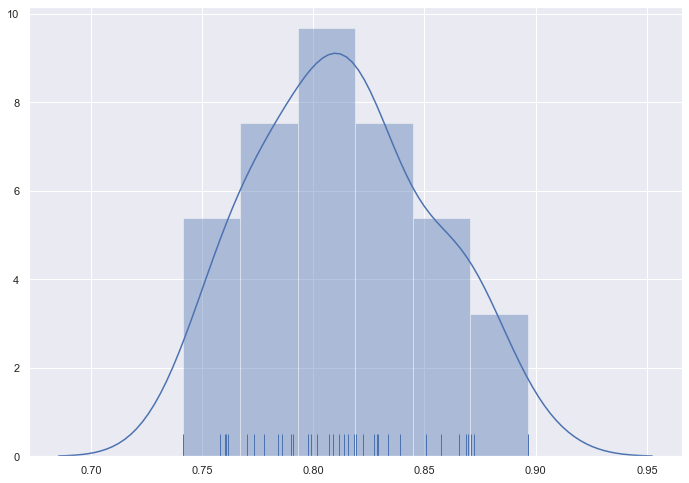}} 
    \end{tabular}
    \caption{$\mathcal{NI}$ values binned by temporal resolution. Scale $j-j+1$ indicates the levels of the temporal wavelet transform bin. The $\{\mbox{min}, \mbox{average}, \mbox{max}\}$ is displayed for each bin. }
    \label{NMI_space} 
\end{figure}
\end{center}

\newpage

\begin{center}
\centering
\begin{figure}[h!]
\centerfloat
    \includegraphics[width=.80\textwidth]{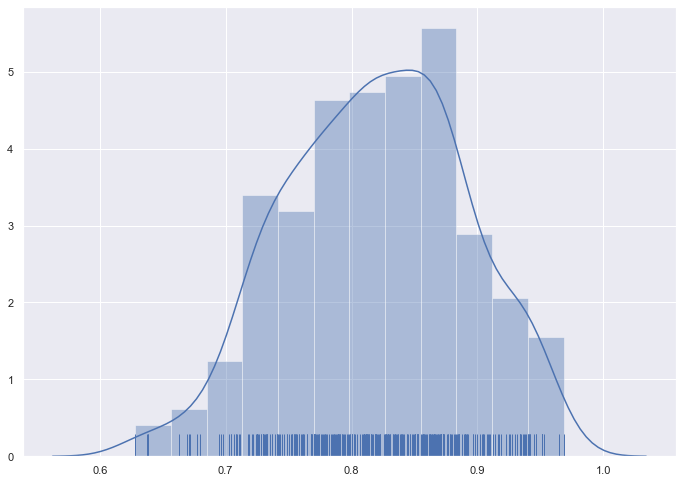}
    \caption{The PDF of adjacent $\mathcal{NI}$ values for CGC with K-means, $K=4, \dots, 20$ on L15 dataset.   $\{\mbox{min}, \mbox{average}, \mbox{max}\} = \{0.63, 0.82, 0.92\}$.}
    \label{NMI_PDF} 
\end{figure}
\end{center}

\subsubsection{L15 MIER algorithm results - creating the coarse-grain ensemble.}

For different $K$ values, the CGC algorithm was run across all the resolutions $\mathcal{L}$ as in Equation \ref{lattice}.  For each fixed $K$, the MIER algorithm was applied to the outputs to discover the reduced ensemble. Once again for a parsimonious exposition, we will only display the MIER algorithm for $K=10$.  

Figure \ref{ER_plots} shows the result of the MIER algorithm for $K = 10$. Figure \ref{ER_plots}a displays the value of $\mathcal{A}(U^{\vec{\ell}})$ for each resolution $\vec{\ell} \in \mathcal{L}$. The value on the vertical access denotes the number of spatial wavelet transforms, while the horizontal axis displays the number of temporal wavelet transforms. Figure \ref{ER_plots}b shows the results of the ratio cut algorithm.  The key resolutions found by running the MIER algorithm are highlighted in a darker shade.

The clusters plotted in Figure \ref{ER_plots}c through Figure \ref{ER_plots}f are the best representative clusterings found by the MIER algorithm. 
Each clustering encapsulates different observed qualitative features from the large ensemble of clusterings $\mathcal{L}$.

\subsection{Discussion}

\subsubsection{CGC Discussion.}

CGC resolution dependence plots in Figure \ref{resolutions} highlight the variability that data resolution introduces into the clustering process. It perhaps comes as little surprise that increasing the number of spatial wavelet transform results in a coarser clustering.  High variance regions, such as the Rocky Mountains, become less resolved as the number of spatial resolutions increases.  However, large structural features such as The Great Plains are persistent across the spatial wavelet coarse-graining.

What is more unexpected is the effect that coarse-graining time has on the clustering. High variability regions remain high variability, however distinctly different clustering patterns do begin to emerge. For instance, how CGC clusters the Northern Rocky Mountains and the Pacific Northwest does seem to depend on the temporal resolution selected. Indeed, in Figure \ref{fig:mr_cluster} we see that increasing temporal scale results adds more ``micro'' biomes to these regions. Low variability regions also depend heavily on the temporal scale.  For example, the North Eastern U.S. splits into more biomes as the temporal scale becomes coarser.  

To better understand which resolutions effect the clustering greatest, we computed the $\mathcal{NI}$ for various adjacent scales. In Table \ref{NMI_single}, we see that there is a large variability in the average $\mathcal{NI}$. This shows that increasing the coarseness at some scale lengths results in a greater loss of information then others. For instance, $\mathcal{NI}(U^{(1,1,1)}, U^{(2,2,1)})$ is large - roughly $90\%$ of the information is retained on average.  This indicates that the difference between clustering at the $(1,1,1)$ resolution ($12$ kilometers) versus the $(2,2,1)$ resolution ($24$ kilometers) isn't significant at the monthly time scale for the L15 dataset.  At higher temporal scales however, it would appear as though the difference between the $12$ kilometers and $24$ kilometer resolution is more consequential. This is likely due to the blurring of the added complexity in regions such as the Northern Rocky Mountains and the Pacific Northwest discussed above.

 Table \ref{NMI_single} allows us to pinpoint individual resolution jumps where large amounts of information was lost. Figures \ref{NMI_time} and \ref{NMI_space} on the other hand, help discover the spatial and temporal scales that are responsible for information loss.  Figures \ref{NMI_time}a (12 to 24 km) and \ref{NMI_time}b (24 to 48 km) demonstrate that regardless of temporal scale and $K$, the expected $\mathcal{NI}$ was approximately the same. However, \ref{NMI_time}c (48km to 96km) is significantly lower, with a maximum value near the mean of the other two scale transitions. This indicates that spatial scale $i=4$ has entered into a regime where the higher frequency features of the L15 dataset have been completely smoothed for more global ones. 
 
 Perhaps more interesting, Figure \ref{NMI_space} allows us to identify the timescales responsible for the largest amount of change. Note that Figures \ref{NMI_space}a (2 to 4 month), \ref{NMI_space}c (8 to 16 months) and \ref{NMI_space}d (16 to 32 months) have similar, right weighted distributions with relatively high expected $\mathcal{NI}$. However, Figures \ref{NMI_space}b (4 to 8 months) and \ref{NMI_space} (32 to 64 months) are different. The transition from 32 to 64 months is significantly more normally distributed with a noticeably lower average $\mathcal{NI}$.  The transition from 4 to 8 months is very different, with its $\mathcal{NI}$ values densely centered near the mean significantly lower than the other four plots. What these plots seem to indicate is that, in terms of clustering the L15 dataset,
is the critical role seasons and many years play.  Indeed, the difference between 2 or 4 months isn't very significant in terms of loss of information.  However, once you start to jump across seasons (Figure \ref{NMI_space}b), new patterns emerge causing a remarkably different clustering. Likewise, the difference between seasons and small number of years  (Figure \ref{NMI_space}b and c) isn't very significant. However, once you start to enter into a larger number of years (Figure \ref{NMI_space}e), climate trends begin to emerge. This climate signal is slower moving compared to the yearly signal, resulting in a noticeably different clustering from an information theoretic standpoint.

\subsubsection{MIER Discussion.}


The MIER algorithm massively reduces the size of the large ensemble $\mathcal{L}$ for each choice of $K$.  In all the experiments run, the size of $\mathcal{L}$ was 24, but the reduced ensemble size is between three and five, with the majority of the cases being four. This illustrates the success of the method in identifying a reduced set of clusterings.  Furthermore, the algorithm is successful at picking resolutions that are sufficiently spaced apart. Consequently, the chosen clusters  accurately represent the dynamical range of all the 24 clusters in the large ensemble. 

Qualitatively, this can be seen by comparing Figures \ref{resolutions} and \ref{ER_plots}.  The six sample plots in Figure \ref{resolutions} are the extreme cases (lowest and highest coarse-graining) and as well as some middle cases. By looking at Figure \ref{ER_plots} we see that, for instance, the cluster $U^{(1,1,1)}$ belongs to component $0$. The representative for component $0$ is the cluster $U^{(2,2,1)}$.  $U^{(1,1,1)}$ on Figure \ref{resolutions} and $U^{(2,2,1)}$ on Figure \ref{ER_plots} are qualitatively similar. Indeed,  $U^{(2,2,1)}$ has structure observed in $U^{(1,1,1)}$ and  $U^{(4,4,1)}$, which is another clustering that belongs to the same connected component. 

\begin{center}
\centering
\begin{figure}[hbt!]
\centerfloat
    \begin{tabular}{cc}
        \subfloat[$\mathcal{A}(U^{\vec{\ell}})$]{\includegraphics[width = 0.45\textwidth]{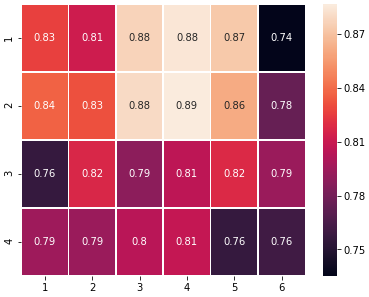}} &
        \subfloat[Ratio-cut with optimal clusters highlighted]{\includegraphics[width = 0.45\textwidth]{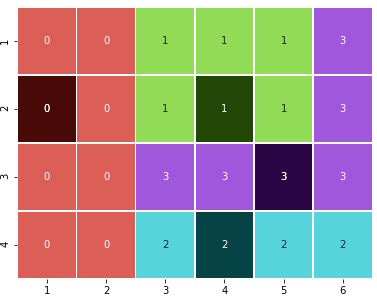}} \\
        
        \subfloat[ $(\ell_1, \ell_2, \ell_3) = (2,2,1)$]{\includegraphics[width = 0.45\textwidth]{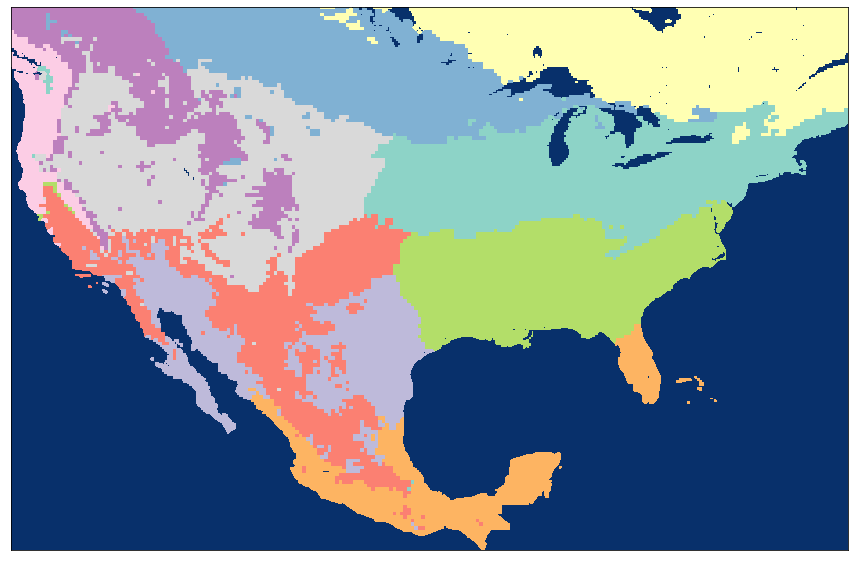}} &
        \subfloat[ $(\ell_1, \ell_2, \ell_3) = (2,2,4)$]{\includegraphics[width = 0.45\textwidth]{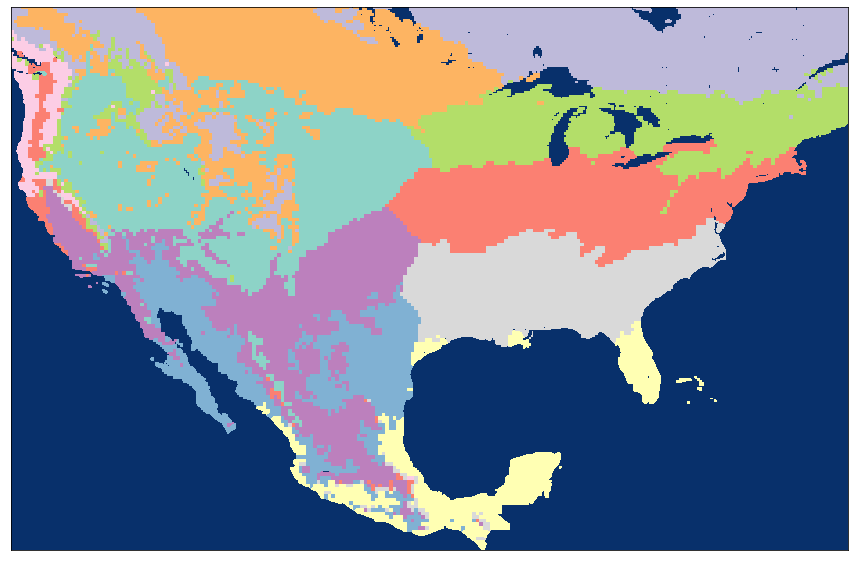}} \\

        \subfloat[ $(\ell_1, \ell_2, \ell_3) = (3,3,5)$]{\includegraphics[width = 0.45\textwidth]{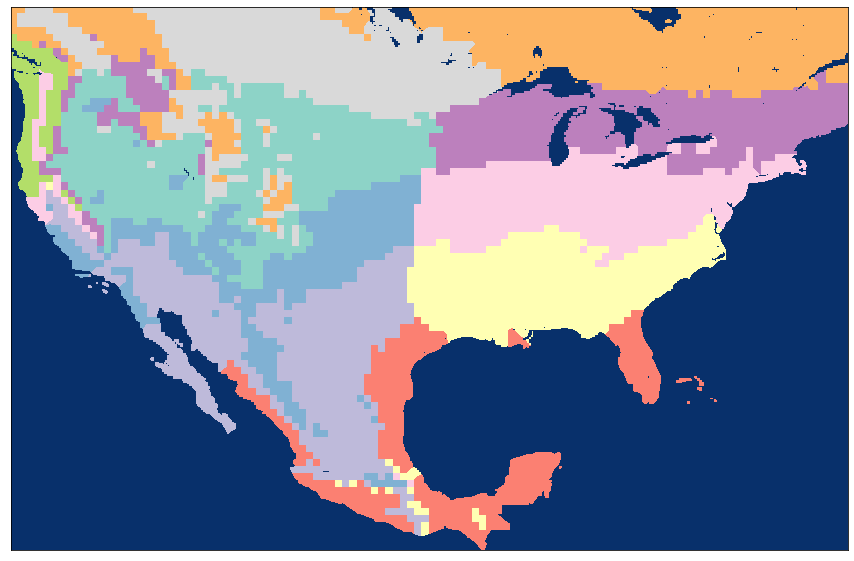}} &
        \subfloat[$(\ell_1, \ell_2, \ell_3) = (4,4,4)$]{\includegraphics[width = 0.45\textwidth]{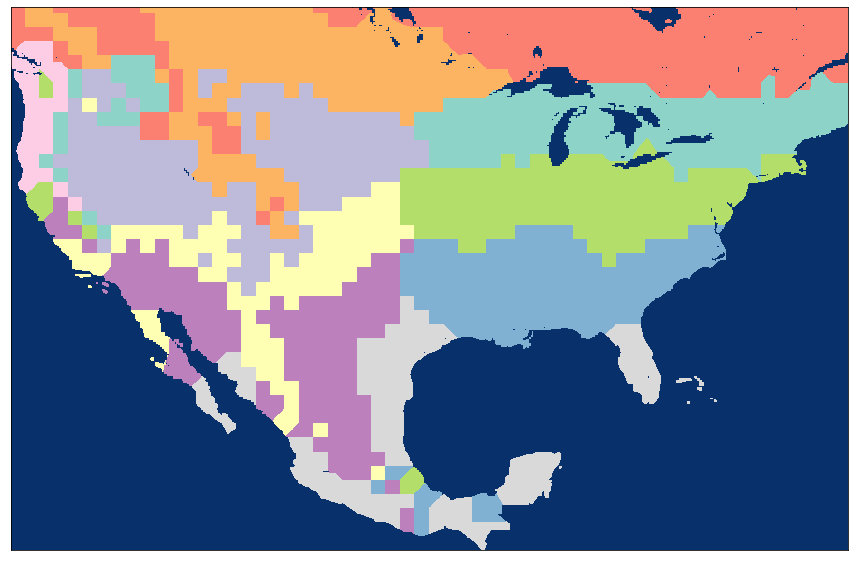}} \\

    \end{tabular}
    \caption{Output of the ER algorithm for $K=10$. a) $\mathcal{A}(U^{\vec{\ell}})$ for each resolution $\vec{\ell} \in \mathcal{L}$, b) results of the ratio cut algorithm and key resolutions, c-f) plots of key resolutions.}
    \label{ER_plots}
\end{figure}
\end{center}

As can be seen from the output of the MIER algorithm, the reduced ensemble can succinctly represent differences across the spatial temporal resolutions. Most of the variance seen between the clusterings at different resolutions is captured within this subset.  From a numerical standpoint, the reduced ensemble is robust as well.

The success of the MIER algorithm can also be identified from an information theoretic standpoint.  As can been seen from Figure \ref{ER_plots}, the expected normalized mutual information between any representative and the other clusters in its component of the graph is usually rather large. Note that while mutual information graph $\mathcal{G}$ is built from all pairs of interactions $\mathcal{NI}(U^{\ell}, U^{\ell'})$, $\ell, \ell' \in \mathcal{L}$, the graph cut is made along adjacent resolutions.  This was common amongst all the runs of the MIER algorithm for $K=4, \dots, 20$. Furthermore, the max value for $\mathcal{A}(U^{\vec{\ell}})$ over each component was almost always greater than the average value obtained by \emph{adjacent} mutual information.  This is noteworthy, since adjacent resolutions will always have a higher mutual information than those spread apart.  This indicates that the MIER algorithm has selected components that were very information theoretically similar.  Thus, picking the best cluster from each component via $\mbox{Argmax}_{\vec{\ell} \in \mathcal{L}_j} \mathcal{A}(U^{\vec{\ell}})$, we have found a small set of clusters  that minimize the information loss across the large ensemble.

\section{Conclusion}


We have shown that scale of data is a non-negligible feature with regards to clustering.  The normalized mutual information drop off at different scale lengths illustrates certain resolutions are producing novel clusters. Consequently, in addition to running several clustering algorithms, it is also important to include several coarse-grain clusterings into your cluster ensemble.  To avoid ballooning the size of the ensemble, its crucial to not consider every possible coarse-graining, but rather a small subset that largely represents every possible resolution.  The MIER algorithm has shown to be a good method to prune the size of the CGC ensemble. This capability to produce an ensemble of classifications representing the diversity of scales provides a direct pathway to better understand clustering sensitivities, illustrating a continued need to assess and mitigate uncertainties resultant from hyperparameter selection.  

As discussed in Section 3.1, the computational complexity to run many CGC's is generally no more expensive than naive K-means.  However, the resultant large ensemble it is more difficult to analyze than a single clustering.  By design, MIER algorithm effectively selects a diverse small subset from the large ensemble. However, as discussed in Figure \ref{fig:consensus}, the additional clusterings from the CGC and MIER framework should be imported into a consensus clustering algorithm.   Further study is needed to assess the confidence across the cluster ensembles within this classification approach.  

\clearpage
\section{Appendix}
\subsection{Continuous Wavelet Transform}

Representing functions via a decomposition into simpler functions is classic. Perhaps the most famous is the Fourier transform and Fourier series.  Recall that given $f \in L^2[0,2\pi]$, the set of square integrable (finite energy) functions  and $\omega \in \mathbb{R}$, the Fourier transform
\[
\hat{f}(\omega) = \frac{1}{2 \pi} \int_0^{2 \pi} f(x) e^{-i \omega x} ~dx
\]
measures the ``amount'' of the frequency $\omega$ the signal $f$ contains. The Fourier series of $f \in L^2[0, 2\pi]$ is a decomposition of $f$ into a (potentially infinite) $L^2-$sum of trigonometric functions.  Concretely, the functions $f_N(x) = \sum_{n=-N}^N \hat{f}(n) e^{inx}$ converge to $f$ in $L^2[0,2\pi]$ \cite{rudin1962fourier}.

The Fourier transform only contains this frequency information, it doesn't know ``where'' this frequency occurs. By taking a window of the domain, one can better localize frequencies. However, there is a Heisenberg's uncertainty type inequality that limits how well one can simultaneously resolve both the frequency and spatial values of a function \cite{folland1997uncertainty}.  The wavelet transform seeks to solve this issue by utilizing a scaling window, which is shifted along the signal.  At each position, the spectrum is calculated.  This process is repeated with varying window lengths, resulting in a collection of time-frequency representations of the signal and multiple resolutions i.e. power at different frequencies and scales. 

Wavelet transforms differ from the Fourier transform in that there is a choice of wavelet function, or kernel, one uses to compute the weights. Rather than integrating against $e^{-i \omega x}$ to measure the amount of frequency $\omega$ the function $f$ has, shifts and dilation's of a ``window'' kernel function are used.  The \textit{mother wavelet} is a function $\Psi \in L^2([a,b])$ that satisfies additional regularity conditions \cite{daubechies1992ten}.  Perhaps the most famous example is Haar wavelet, given by
\[
\Psi(x) = \left\{ 
\begin{array}{cc}
   1  &  0 \leq x < \frac{1}{2}\\
  -1   &  \frac{1}{2} \leq x < 1\\
  0 & \mbox{else}.
\end{array}
\right.
\]
Wavelets are generated by scaling and translated the mother wavelet.  Given a scale factor $s$ and translation factor $\tau$, we denote the wavelet scaled by $s$ and shifted by $\tau$ as
\[
\Psi_{s,\tau}(x) = \frac{1}{\sqrt{s}} \Psi\left( \frac{x - \tau}{s} \right).
\]
Just as the Fourier transform at $\omega$ measures the amount of frequency $\Omega$ in $f$ by convolution with $e^{-i \omega x}$, the wavelet transform measure the ``amount'' of the signal at scale $s$ and translation $\tau$ by convolution.  The (continuous) wavelet transform of $f$ at $s, \tau$ is
\[
\hat{f}(s,\tau) = \int_0^{2\pi} f(x) \overline{\Psi_{s,\tau}(x) } ~dx
\]
where the bar denotes complex conjugate.  In our work, we will only be working with real valued wavelet functions, so the conjugate is not necessary.  Thus, the wavelet transform $\hat{f}(s,\tau)$ will always be real valued. 

\subsection{Discrete Wavelets}

By taking discrete steps in the frequency space, the Fourier transform was leveraged to create the Fourier series. Similarly, by fixing a scale and translation and taking integral steps, one can create a wavelet series representation of $f$.  Usually, one fixes $s=2$, and $\tau=1$ (so-called dyadic sampling of scale/translation space), and modifies the wavelet transform to
\[
\Psi_{\ell,k}(x) = \frac{1}{\sqrt{2^\ell}} \Psi\left( \frac{x - 2^\ell}{2^\ell} \right),
\]
where $\ell, k \in \mathbb{Z}$.  The $\Psi_{\ell,k}$ are known as the discrete wavelets. If the discrete wavelets satisfy a very natural bounding condition called the \textit{frame bounds}, then the initial signal can be reconstructed perfectly with (potentially infinite) $L^2-$sum of $\Psi_{\ell,k}$  \cite{daubechies1992ten}. That is,
\[
f = \sum_{\ell=-\infty}^\infty \sum_{k=-\infty}^\infty \hat{f}(\ell,k) \Psi_{\ell,k}
\]
where convergence is in $L^2$.

\subsection{Discrete Wavelet Transform - Wavelet Filter Bank}

Clearly in practice, one cannot compute the infinite number of $\hat{f}(\ell,k)$.  Luckily in practice, one wouldn't have to anyways. At any given scale, the number of translations is bounded by the length of the signal interval.  Therefore, it suffices to limit the range of scales one needs to compute.  From Fourier analysis, we know that time compression by factor of $2$ will stretch and shift the frequency spectrum by a factor of $2$.  If, for example, $\Psi_{1,0}$ covers the upper bound of the frequency spectrum of $f$, then further dilations will begin to cover the whole frequency spectrum of $f$.  See Figure \ref{fig:frequency}. 

\begin{center}
\centering
\begin{figure}[h!]
\centerfloat
    \includegraphics[width = .8\textwidth]{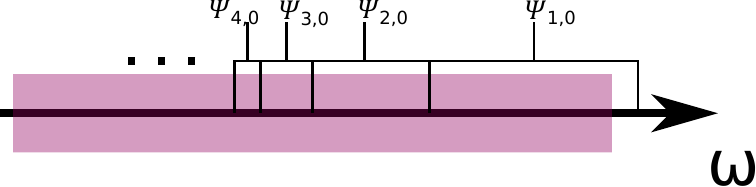}
    \caption{Diagram demonstrating how the different dilations cover the frequency spectrum of $f$.  The bar in red indicates the frequency domain of $f$, and the boxes for each $\psi_{\ell,0}$ show what portion of the frequency spectrum the wavelet captures.}
    \label{fig:frequency}
\end{figure}
\end{center}
 Because the window width for the frequency spectrum is halved at each iteration, it isn't possible to cover zero with a finite number of iterations. For this reason, after a finite number of steps, the residual low passed signal is collected.  The low frequency function is called the \textit{scaling function}, denoted $\Phi$. For the Haar wavelet, 
 \[
\Phi(x) = \left\{ 
\begin{array}{cc}
   1  &  0 \leq x < 1\\
  0 & \mbox{else}.
\end{array}
\right.
\]
 This process of splitting the frequency domain via low-pass and band-pass filters is an example of a \textit{filter bank}  \cite{mallat1989theory}. 
 
\begin{figure}[h!]
    \centering
    \includegraphics[width = .9\textwidth]{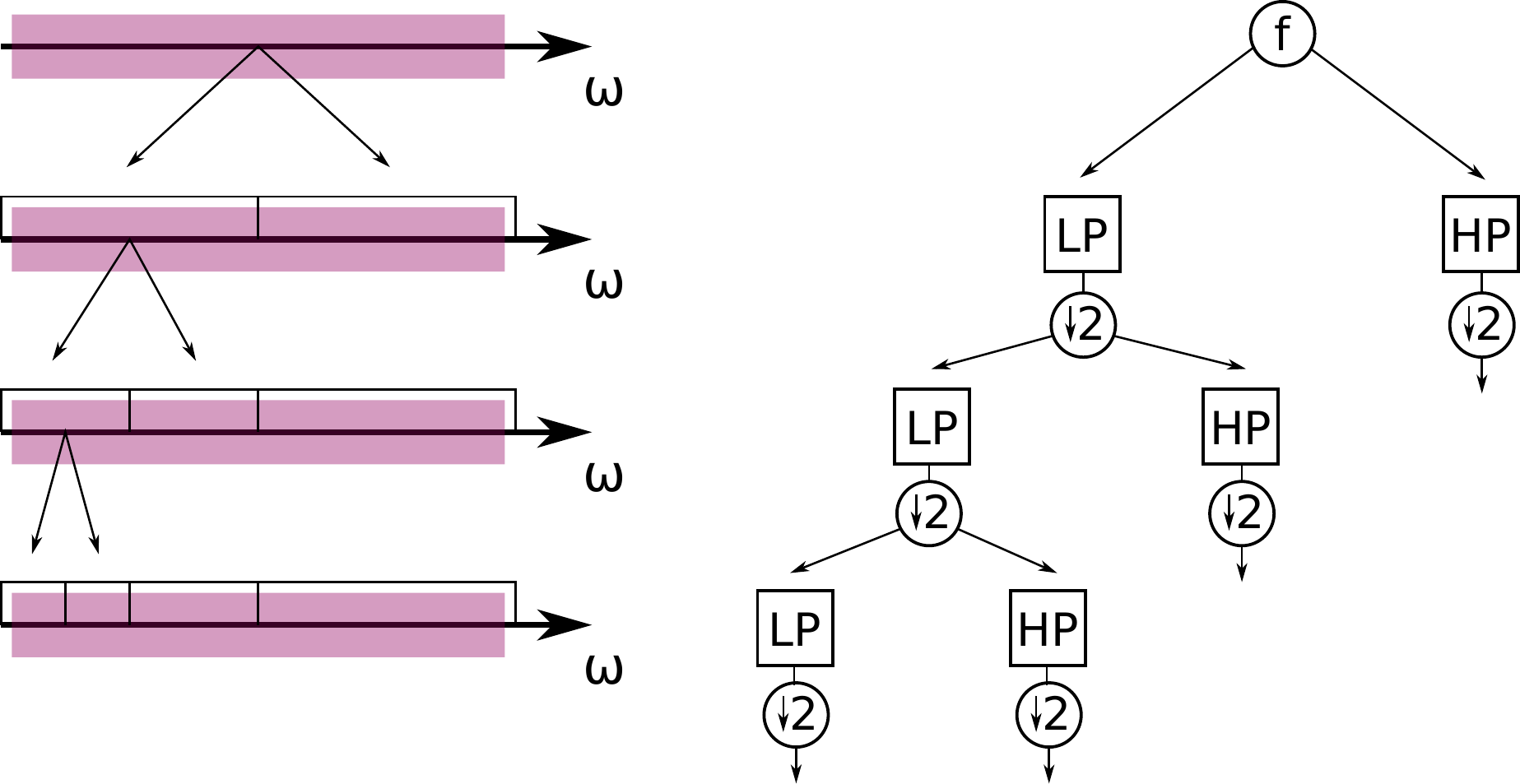}
    \caption{Diagram of a 3-layer wavelet filter bank for a signal $f$.  The frequency domain of $f$ is split each iteration via a low pass/high pass filter.  After each filter, the resultant signal is down-sampled.  }
    \label{fig:filter_bank}
\end{figure}

 In practice, the filter bank is created by selecting a scale level $\ell$, and iteratively computing the wavelet coefficients up to the chosen scale.  For example, if $\ell=2$,  convolution with the  wavelets $\Psi_{1,0}$, $\Psi_{1,1}$ will cover the top half of the frequency spectrum for $f$ providing a high pass filter.  Taking the scaling function from the first iteration, convolution with $\Psi_{2,0}, \Psi_{2,1}, \Psi_{2,2}, \Psi_{2,3}$ will cover the next quarter of the frequency spectrum for $f$.  The residual is the low passed signal. See Figure \ref{fig:filter_bank}.

A computationally efficient implementation of this filter bank method is easily implemented for a discrete 1-D signal $f$ of length $N$.  Here, the convolution and downsampling can be encoded into a single vector-matrix multiplication $g = H f$.  The output $g = [g_l | g_h]$ will be a vector approximately the same length as $f$.  Two vectors $g_l$ and $g_h$ that make up $g$ consist of the low-pass and high-pass portions of $f$ respectively \cite{thyagarajan2011discrete}.

\begin{figure}[h!]
    \centering
    \includegraphics[scale=.7]{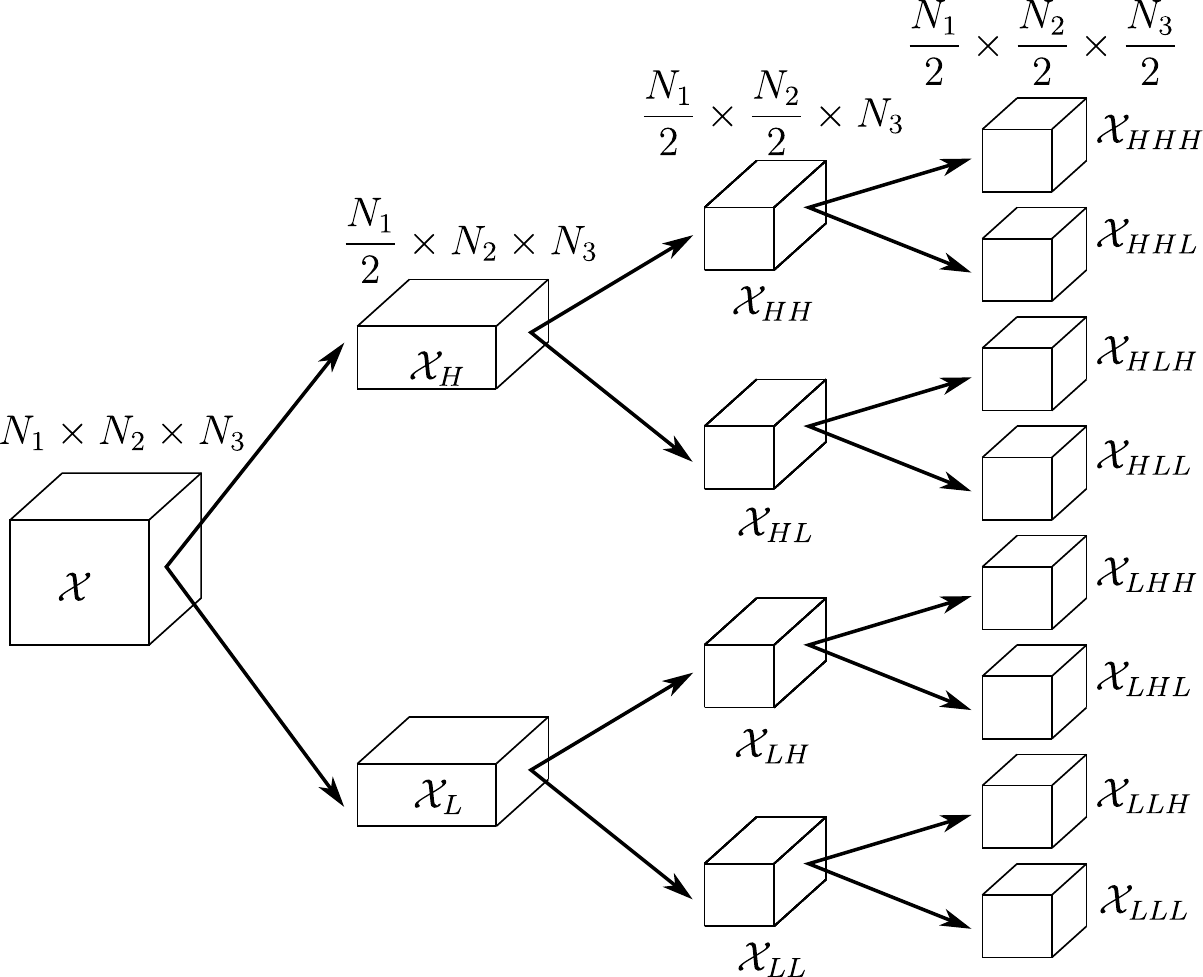}
    \caption{Diagram of (separable) DWT applied to a 3-tensor of data.  After each application of a 1-D wavelet transform, the size of the tensor is reduced by a factor of two. }
    \label{fig:3dwavelet}
\end{figure}

For multidimensional signals such as a tensor, there are several ways to compute the discrete wavelet transform.  The most popular is the separable method, whereby the 1-D DWT is applied sequentially along each axis. For example, given a 3-way tensor data of size $N_1 \times N_2 \times N_3$, the 1-D DWT filterbank is first applied along axis 1.  This splits the data into two chunks of size roughly $\frac{N_1}{2} \times N_2 \times N_3$. The 1-D DWT filterbank can then be applied again to the next axis, again reducing the size of the data \cite{chun2010tutorial}. See figure \ref{fig:3dwavelet} for a diagrammatic explanation.

The complexity of the 1-D DWT can be shown to be $\mathcal{O}(N)$ \cite{shukla2013efficient}.  The complexity for higher-order wavelet transforms will depend on the choice of wavelets, but a worst-case upper bound for a separable DWT of a third-order tensor is $\mathcal{O}(N_1 N_2 N_3)$.  For a comprehensive overview of wavelets, see \cite{jensen2001ripples}.

\subsection{Example of Discrete Wavelet Transform}

Consider the 1-D signal $f=[1,2,3,4,5,6,7,8]$. Suppose we want to compute the third-level Haar wavelet transform of $f$. The matrix for the Haar wavelet transform is given by

\[
H = \frac{1}{\sqrt{8}}
\begin{pmatrix}
1 & 1 & 1 & 1 & 1 & 1 & 1 & 1 \\
1 & 1 & 1 & 1 & -1 & -1 & -1 & -1 \\
\sqrt{2} & \sqrt{2} & -\sqrt{2} & -\sqrt{2} & 0 & 0 & 0 & 0 \\
0 & 0 & 0 & 0 & \sqrt{2} & \sqrt{2} & -\sqrt{2} & -\sqrt{2}\\
2 & -2 & 0 & 0 & 0 & 0 & 0 & 0 \\
0 & 0 & 2 & -2 & 0 & 0 & 0 & 0 \\
0 & 0 & 0 & 0 & 2 & -2 & 0 & 0 \\
0 & 0 & 0 & 0 & 0 & 0 & 2 & -2 
\end{pmatrix}
\begin{matrix}
\Phi \\
\Psi_1 \\
\Psi_{2,0} \\
\Psi_{2,1} \\
\Psi_{3,0} \\
\Psi_{3,1} \\
\Psi_{3,2} \\
\Psi_{3,3}
\end{matrix}
\]

Here, we have written the $\Phi$ and $\Psi_{\ell, k}$ next to the rows that the represent.  In order to compute the DWT, we compute $Hf$.  Doing the matrix multiplication yields
\[
H f = \frac{1}{\sqrt{8}} \begin{pmatrix}
36 \\
\hline
-16\\
\hline
-2^\frac{3}{2}\\
-2^\frac{3}{2}\\
\hline
-2\\
-2\\
-2\\
-2\\
\end{pmatrix}
\]
where we have added separation lines to indicate the different detail coefficients from the different coarse-grainings. The top level is the approximation third level's approximation coefficient.  

\section*{Acknowledgements}

This research was supported as part of the Energy Exascale Earth System Model
(E3SM) project, funded by the U.S. Department of Energy, Office of Science,
Office of Biological and Environmental Research as well as LANL laboratory directed research and development (LDRD) grant 20190020DR.
High-performance computing time was conducted at 
Los Alamos Nat. Lab. Institutional Computing, US DOE NNSA (DE-AC52-06NA25396).

\newpage
\section*{References}

 \bibliographystyle{plain}
 \bibliography{bib}

\begin{thebibliography}{10}

\bibitem{alexandrov2019nonnegative}
Boian~S Alexandrov, Valentin~G Stanev, Velimir~V Vesselinov, and Kim~{\O}
  Rasmussen.
\newblock Nonnegative tensor decomposition with custom clustering for
  microphase separation of block copolymers.
\newblock {\em Statistical Analysis and Data Mining: The ASA Data Science
  Journal}, 12(4):302--310, 2019.

\bibitem{alexandrov2014blind}
Boian~S Alexandrov and Velimir~V Vesselinov.
\newblock Blind source separation for groundwater pressure analysis based on
  nonnegative matrix factorization.
\newblock {\em Water Resources Research}, 50(9):7332--7347, 2014.

\bibitem{amodei2016concrete}
Dario Amodei, Chris Olah, Jacob Steinhardt, Paul Christiano, John Schulman, and
  Dan Man{\'e}.
\newblock Concrete problems in ai safety.
\newblock {\em arXiv preprint arXiv:1606.06565}, 2016.

\bibitem{bishop2006pattern}
Christopher~M Bishop.
\newblock {\em Pattern recognition and machine learning}.
\newblock Springer Science+ Business Media, 2006.

\bibitem{cao2013robust}
Xiaochun Cao, Xingxing Wei, Yahong Han, Yi~Yang, and Dongdai Lin.
\newblock Robust tensor clustering with non-greedy maximization.
\newblock In {\em Twenty-Third International Joint Conference on Artificial
  Intelligence}, 2013.

\bibitem{caruana2004ensemble}
Rich Caruana, Alexandru Niculescu-Mizil, Geoff Crew, and Alex Ksikes.
\newblock Ensemble selection from libraries of models.
\newblock In {\em Proceedings of the twenty-first international conference on
  Machine learning}, page~18, 2004.

\bibitem{chun2010tutorial}
Liu Chun-Lin.
\newblock A tutorial of the wavelet transform.
\newblock {\em NTUEE, Taiwan}, 2010.

\bibitem{cichocki2009nonnegative}
Andrzej Cichocki, Rafal Zdunek, Anh~Huy Phan, and Shun-ichi Amari.
\newblock {\em Nonnegative matrix and tensor factorizations: applications to
  exploratory multi-way data analysis and blind source separation}.
\newblock John Wiley \& Sons, 2009.

\bibitem{daubechies1992ten}
Ingrid Daubechies.
\newblock {\em Ten lectures on wavelets}, volume~61.
\newblock Siam, 1992.

\bibitem{ding2008equivalence}
Chris Ding, Xiaofeng He, Horst~D Simon, and Rong Jin.
\newblock On the equivalence of nonnegative matrix factorization and
  k-means-spectral clustering.
\newblock 2008.

\bibitem{dom2002information}
Byron~E Dom.
\newblock An information-theoretic external cluster-validity measure.
\newblock In {\em Proceedings of the Eighteenth conference on Uncertainty in
  artificial intelligence}, pages 137--145. Morgan Kaufmann Publishers Inc.,
  2002.

\bibitem{fahad2014survey}
Adil Fahad, Najlaa Alshatri, Zahir Tari, Abdullah Alamri, Ibrahim Khalil,
  Albert~Y Zomaya, Sebti Foufou, and Abdelaziz Bouras.
\newblock A survey of clustering algorithms for big data: Taxonomy and
  empirical analysis.
\newblock {\em IEEE transactions on emerging topics in computing},
  2(3):267--279, 2014.

\bibitem{fern2008cluster}
Xiaoli~Z Fern and Wei Lin.
\newblock Cluster ensemble selection.
\newblock {\em Statistical Analysis and Data Mining: The ASA Data Science
  Journal}, 1(3):128--141, 2008.

\bibitem{folland1997uncertainty}
Gerald~B Folland and Alladi Sitaram.
\newblock The uncertainty principle: a mathematical survey.
\newblock {\em Journal of Fourier analysis and applications}, 3(3):207--238,
  1997.

\bibitem{hadjitodorov2006moderate}
Stefan~T Hadjitodorov, Ludmila~I Kuncheva, and Ludmila~P Todorova.
\newblock Moderate diversity for better cluster ensembles.
\newblock {\em Information Fusion}, 7(3):264--275, 2006.

\bibitem{huang2008simultaneous}
Heng Huang, Chris Ding, Dijun Luo, and Tao Li.
\newblock Simultaneous tensor subspace selection and clustering: the
  equivalence of high order svd and k-means clustering.
\newblock In {\em Proceedings of the 14th ACM SIGKDD international conference
  on Knowledge Discovery and Data mining}, pages 327--335, 2008.

\bibitem{jegelka2009approximation}
Stefanie Jegelka, Suvrit Sra, and Arindam Banerjee.
\newblock Approximation algorithms for tensor clustering.
\newblock In {\em International Conference on Algorithmic Learning Theory},
  pages 368--383. Springer, 2009.

\bibitem{jensen2001ripples}
Arne Jensen and Anders la~Cour-Harbo.
\newblock {\em Ripples in mathematics: the discrete wavelet transform}.
\newblock Springer Science \& Business Media, 2001.

\bibitem{kolda2009tensor}
Tamara~G Kolda and Brett~W Bader.
\newblock Tensor decompositions and applications.
\newblock {\em SIAM review}, 51(3):455--500, 2009.

\bibitem{kottek2006world}
Markus Kottek, J{\"u}rgen Grieser, Christoph Beck, Bruno Rudolf, and Franz
  Rubel.
\newblock World map of the k{\"o}ppen-geiger climate classification updated.
\newblock {\em Meteorologische Zeitschrift}, 15(3):259--263, 2006.

\bibitem{kuncheva2004using}
Ludmila~I Kuncheva and Stefan~Todorov Hadjitodorov.
\newblock Using diversity in cluster ensembles.
\newblock In {\em 2004 IEEE International Conference on Systems, Man and
  Cybernetics (IEEE Cat. No. 04CH37583)}, volume~2, pages 1214--1219. IEEE,
  2004.

\bibitem{lee1999learning}
Daniel~D Lee and H~Sebastian Seung.
\newblock Learning the parts of objects by non-negative matrix factorization.
\newblock {\em Nature}, 401(6755):788--791, 1999.

\bibitem{livneh2015spatially}
Ben Livneh, Theodore~J Bohn, David~W Pierce, Francisco Munoz-Arriola, Bart
  Nijssen, Russell Vose, Daniel~R Cayan, and Levi Brekke.
\newblock A spatially comprehensive, hydrometeorological data set for mexico,
  the us, and southern canada 1950--2013.
\newblock {\em Scientific data}, 2:150042, 2015.

\bibitem{lopez2019unsupervised}
Cesar~A Lopez, Velimir~V Vesselinov, Sandrasegaram Gnanakaran, and Boian~S
  Alexandrov.
\newblock Unsupervised machine learning for analysis of phase separation in
  ternary lipid mixture.
\newblock {\em Journal of chemical theory and computation}, 15(11):6343--6357,
  2019.

\bibitem{lu2016statistical}
Yu~Lu and Harrison~H Zhou.
\newblock Statistical and computational guarantees of lloyd's algorithm and its
  variants.
\newblock {\em arXiv preprint arXiv:1612.02099}, 2016.

\bibitem{mahajan2009planar}
Meena Mahajan, Prajakta Nimbhorkar, and Kasturi Varadarajan.
\newblock The planar k-means problem is np-hard.
\newblock In {\em International Workshop on Algorithms and Computation}, pages
  274--285. Springer, 2009.

\bibitem{mallat1989theory}
Stephane~G Mallat.
\newblock A theory for multiresolution signal decomposition: the wavelet
  representation.
\newblock {\em IEEE transactions on pattern analysis and machine intelligence},
  11(7):674--693, 1989.

\bibitem{netzel2016using}
Pawel Netzel and Tomasz Stepinski.
\newblock On using a clustering approach for global climate classification.
\newblock {\em Journal of Climate}, 29(9):3387--3401, 2016.

\bibitem{ng2002spectral}
Andrew~Y Ng, Michael~I Jordan, and Yair Weiss.
\newblock On spectral clustering: Analysis and an algorithm.
\newblock In {\em Advances in neural information processing systems}, pages
  849--856, 2002.

\bibitem{nguyen2007consensus}
Nam Nguyen and Rich Caruana.
\newblock Consensus clusterings.
\newblock In {\em Seventh IEEE International Conference on Data Mining (ICDM
  2007)}, pages 607--612. IEEE, 2007.

\bibitem{rudin1962fourier}
Walter Rudin.
\newblock {\em Fourier analysis on groups}, volume 121967.
\newblock Wiley Online Library, 1962.

\bibitem{schein2016bayesian}
Aaron Schein, Mingyuan Zhou, David~M Blei, and Hanna Wallach.
\newblock Bayesian poisson tucker decomposition for learning the structure of
  international relations.
\newblock {\em arXiv preprint arXiv:1606.01855}, 2016.

\bibitem{shukla2013efficient}
Kaushal~K Shukla and Arvind~K Tiwari.
\newblock {\em Efficient algorithms for discrete wavelet transform: with
  applications to denoising and fuzzy inference systems}.
\newblock Springer Science \& Business Media, 2013.

\bibitem{stanev2018unsupervised}
Valentin Stanev, Velimir~V Vesselinov, A~Gilad Kusne, Graham Antoszewski,
  Ichiro Takeuchi, and Boian~S Alexandrov.
\newblock Unsupervised phase mapping of x-ray diffraction data by nonnegative
  matrix factorization integrated with custom clustering.
\newblock {\em npj Computational Materials}, 4(1):1--10, 2018.

\bibitem{thornthwaite1948approach}
Charles~W Thornthwaite et~al.
\newblock An approach toward a rational classification of climate.
\newblock {\em Geographical review}, 38(1):55--94, 1948.

\bibitem{thyagarajan2011discrete}
KS~Thyagarajan.
\newblock Discrete wavelet transform.
\newblock 2011.

\bibitem{vesselinov2019unsupervised}
Velimir~V Vesselinov, Maruti~Kumar Mudunuru, Satish Karra, D~O'Malley, and
  Boian~S Alexandrov.
\newblock Unsupervised machine learning based on non-negative tensor
  factorization for analyzing reactive-mixing.
\newblock {\em Journal of Computational Physics}, 395:85--104, 2019.

\bibitem{vinh2010information}
Nguyen~Xuan Vinh, Julien Epps, and James Bailey.
\newblock Information theoretic measures for clusterings comparison: Variants,
  properties, normalization and correction for chance.
\newblock {\em Journal of Machine Learning Research}, 11(Oct):2837--2854, 2010.

\bibitem{von2007tutorial}
Ulrike Von~Luxburg.
\newblock A tutorial on spectral clustering.
\newblock {\em Statistics and computing}, 17(4):395--416, 2007.

\bibitem{wagner1993between}
Dorothea Wagner and Frank Wagner.
\newblock Between min cut and graph bisection.
\newblock In {\em International Symposium on Mathematical Foundations of
  Computer Science}, pages 744--750. Springer, 1993.

\bibitem{zhang2009parallel}
Qiang Zhang, Michael~W Berry, Brian~T Lamb, and Tabitha Samuel.
\newblock A parallel nonnegative tensor factorization algorithm for mining
  global climate data.
\newblock In {\em International Conference on Computational Science}, pages
  405--415. Springer, 2009.

\bibitem{zscheischler2012climate}
Jakob Zscheischler, Miguel~D Mahecha, and Stefan Harmeling.
\newblock Climate classifications: the value of unsupervised clustering.
\newblock {\em Procedia Computer Science}, 9:897--906, 2012.

\end{thebibliography}
 \end{document}